\documentclass{article}

    \PassOptionsToPackage{numbers, compress}{natbib}

\usepackage[preprint]{neurips_2025}
\usepackage{amsmath}
\usepackage{mathrsfs}
\usepackage{algorithm}
\usepackage{algorithmic}
\usepackage{subcaption} 
\usepackage{graphicx}
\newtheorem{theorem}{Theorem}[section] 
\newtheorem{definition}{Definition}[section] 
\newtheorem{lemma}[theorem]{Lemma}

\newtheorem{assumption}{Assumption}[section]
\usepackage{physics}
\newcommand{\innerprod}[2]{\langle #1, #2 \rangle}

\usepackage[utf8]{inputenc} 
\usepackage[T1]{fontenc}    
\usepackage{hyperref}       
\usepackage{url}            
\usepackage{booktabs}       
\usepackage{amsfonts}       
\usepackage{nicefrac}       
\usepackage{microtype}      
\usepackage{xcolor}         

\title{Policy Newton Algorithm in Reproducing Kernel Hilbert Space}

\author{
  Yixian Zhang, Huaze Tang, Chao Wang, Wenbo Ding \\
  Tsinghua University
}

\begin{document}

\maketitle

\begin{abstract}
Reinforcement learning (RL) policies represented in Reproducing Kernel Hilbert Spaces (RKHS) offer powerful representational capabilities. While second-order optimization methods like Newton's method demonstrate faster convergence than first-order approaches, current RKHS-based policy optimization remains constrained to first-order techniques. This limitation stems primarily from the intractability of explicitly computing and inverting the infinite-dimensional Hessian operator in RKHS. We introduce Policy Newton in RKHS, the first second-order optimization framework specifically designed for RL policies represented in RKHS. Our approach circumvents direct computation of the inverse Hessian operator by optimizing a cubic regularized auxiliary objective function. Crucially, we leverage the Representer Theorem to transform this infinite-dimensional optimization into an equivalent, computationally tractable finite-dimensional problem whose dimensionality scales with the trajectory data volume. We establish theoretical guarantees proving convergence to a local optimum with a local quadratic convergence rate. Empirical evaluations on a toy financial asset allocation problem validate these theoretical properties, while experiments on standard RL benchmarks demonstrate that Policy Newton in RKHS achieves superior convergence speed and higher episodic rewards compared to established first-order RKHS approaches and parametric second-order methods. Our work bridges a critical gap between non-parametric policy representations and second-order optimization methods in reinforcement learning.
\end{abstract}

\section{Introduction}
Representing policies within Reproducing Kernel Hilbert Spaces (RKHS) offers a powerful non-parametric alternative in reinforcement learning (RL), valued for its representational flexibility, potential for improved sample efficiency, and capacity for dynamic adjustment during learning \citep{RKHS_pro, meta_RKHS, distribution_RL}. This approach has demonstrated success in various RL domains, including meta-RL \citep{meta_RKHS} and distributional RL \citep{distribution_RL}. Despite these representational advantages, optimization methods for RKHS policies have remained primarily limited to first-order approaches. The RKHS Policy Gradient \citep{RKHS_TAC}, which achieves policy updates by adding gradient-derived functions in RKHS, represents the current standard. However, this approach inherits the fundamental convergence limitations common to all first-order methods - namely slow convergence in complex optimization landscapes characterized by high curvature or narrow valleys.

In parametric policy representations, second-order optimization methods have emerged as effective solutions to these convergence challenges. While first-order methods like Policy Gradient \citep{PG1} are widely implemented due to their simplicity, they often exhibit slow convergence and sensitivity to the optimization landscape's curvature, particularly when dealing with ill-conditioned problems \citep{Newton2016}. Second-order methods, exemplified by the Policy Newton algorithm \citep{Newton1_policy, Newton_policy2}, address these limitations by incorporating Hessian curvature information, enabling potentially faster convergence rates and more appropriately scaled updates. These advantages make second-order methods particularly compelling candidates for accelerating learning in RKHS policy optimization.

The natural progression towards faster optimization—developing a Policy Newton method directly within the RKHS—poses significant theoretical and practical challenges. Unlike the finite-dimensional case, the Hessian analogue in RKHS corresponds to the second-order Fréchet derivative of the expected cumulative reward function. This derivative is an operator acting on the function space, and computing its inverse explicitly, as required by standard Newton methods, is generally intractable in this infinite-dimensional setting. Existing research on second-order methods in RKHS has primarily focused on regret bounds in online learning settings with specific distributed data \citep{RKHS_Newton1, RKHS_Newton2, RKHS_Newton3, RKHS_Newton4}, leaving a critical gap for policy optimization in RL where the data distribution shifts with the policy.

To bridge this gap, this paper introduces the \textbf{Policy Newton in RKHS} algorithm, the first second-order optimization framework specifically tailored for policies represented within RKHS in the RL context. Our approach circumvents the explicit computation of the infinite-dimensional Hessian operator in policy optimization by reformulating the problem through a cubic regularized auxiliary objective function within the RKHS \citep{Cubic_newton, regularization_Newton}. Crucially, we leverage the Representer Theorem \citep{RepresenterTheorem} to demonstrate that this infinite-dimensional optimization problem is equivalent to solving a finite-dimensional optimization problem in Euclidean space, whose dimension scales with the amount of trajectory data used in the estimate. This makes the approach computationally feasible.

Our main contributions are summarized as follows:
\begin{itemize}
\item We propose the first second-order optimization algorithm for policy in RKHS, comprised of two key components: (1) We derive the second-order Fréchet derivative as the Hessian operator and introduce a cubic regularized auxiliary function to find the update step, avoiding the need to compute the intractable inverse operator; (2) We reformulate the infinite-dimensional optimization problem into an equivalent finite-dimensional problem in Euclidean space using the Representer Theorem, making the approach computationally tractable.

\item We establish theoretical guarantees for the proposed algorithm, proving convergence to a local optimum, and demonstrating a quadratic convergence rate. Our empirical evaluations on a toy problem verify these theoretical properties and show that Policy Newton in RKHS achieves superior performance in terms of episodic reward compared to baseline methods, with an enhanced ability to escape local optima.
\end{itemize}

\section{Preliminaries}
\subsection{Policy Newton in reinforcement learning}
In reinforcement learning, a Markov decision process is defined by the tuple $(\mathcal{S}, \mathcal{A}, P, r, \gamma, \rho)$ where $\mathcal{S}$ denotes the state space; $\mathcal{A}$ denotes the action space; $P(s_{t+1} \mid s_t, a_t)$ represents the transition probability function; $r(s_t, a_t)$ is the reward function; $\gamma \in [0,1)$ is the discount factor; and $\rho(s_0)$ is the initial state distribution. Actions are selected according to a policy $\pi(a_t \mid s_t)$, which defines a probability distribution over actions conditional on the current state. A trajectory is denoted by $\tau = (s_0, a_0, \ldots, a_{T-1}, s_T)$, where $s_0 \sim \rho(s_0)$ and $T$ is the episode length. We denote the probability of trajectory $\tau$ following a policy $\pi$ as $p(\tau ; \pi)$.
The objective of RL is to minimize over $\pi$ the expected discounted cumulative reward given by $J(\pi)=\mathbb{E}_{\tau \sim p(\tau ; \pi)}\left[\sum_{t=0}^{T-1} \gamma^{t-1} r\left(s_{t}, a_{t}\right)\right]$, 
where $\gamma$ is the discount factor. Typically, the policy is parameterized by a vector $\theta \in \mathbb{R}^{d}$ and the notation $\pi_{\theta}$ is used as a shorthand for the distribution $\pi\left(a_{t} \mid s_{t} ; \theta\right)$. 
The target of RL is to find a parameter \citep{Suttonbook}
\begin{equation}
\nonumber
\theta^{*} = \underset{\theta \in \mathbb{R}^{d}}{\operatorname{argmin}} J(\pi_{\theta}) . 
\end{equation}
The Policy Gradient method \citep{Cubic_newton} is utilized to find the optimal parameter using the gradient $\nabla_{\theta} J(\pi_{\theta})$ of the expected reward $J(\pi_{\theta})$:

\begin{equation}
\nonumber
    \nabla_{\theta} J(\pi_{\theta}) = \mathbb{E}_{\tau \sim p(\tau ; \pi)}\left[ \sum_{t=0}^{T-1} \Psi_{t}(\tau) \nabla_{\theta} \log \pi\left(a_{t} \mid s_{t} ; \theta\right)\right],
\end{equation}
where $\Psi_{t}(\tau) = \sum_{i=t}^{T-1} \gamma^{i-1} r\left(s_{i}, a_{i}\right)$ denotes the cumulative reward starting from $(s_t,a_t)$ in trajectory $\tau$. The Policy Gradient has enjoyed success in many fields, but it is not scale invariant and the search direction is often poorly-scaled \citep{Newton2016}. To accelerate the optimization, the second-order information is integrated in the Policy Newton method by using the Hessian $\nabla_{\theta}^{2} J(\pi_{\theta})$ \citep{Newton2019}:
\begin{equation}
\label{Policy_Newton_formula}
\begin{aligned}
        \nabla_{\theta}^{2} J(\pi_{\theta}) = &\mathbb{E}_{\tau \sim p(\tau ; \pi_{\theta})}\left[\sum_{t=0}^{T-1} \Psi_{t}(\tau) \nabla_{\theta} \log \pi\left(a_{t} \mid s_{t} ; \theta\right)
            \times \sum_{t^{\prime}=0}^{T-1} \nabla_{\theta}^{\top} \log \pi\left(a_{t^{\prime}} \mid s_{t^{\prime}} ; \theta\right) \right. \\
       &\left. + \sum_{t=0}^{T-1} \Psi_{t}(\tau) \nabla_{\theta}^{2} \log \pi\left(a_{t} \mid s_{t} ; \theta\right) \right] = \mathbb{E}_{\tau \sim p(\tau ; \pi_{\theta})}\left[ H_{\theta}(\tau ; \pi_{\theta}) \right],
\end{aligned}
\end{equation}
where $H_{\theta}(\tau ; \pi)$ represents the Hessian matrix.
During the training of the RL, the direct calculation of the gradient is infeasible. Therefore, in the $k$-th iteration of the training, the gradient $\nabla_{\theta} J(\theta_k)$ is estimated among the sampled trajectory set $\mathcal{T}$ \citep{PG1}:
\begin{equation}
\label{Est_gradient}
\nabla_{\theta} \hat{J}(\theta_k) = \frac{1}{N} \sum_{\tau \in \mathcal{T}_{N}} \sum_{t=0}^{T-1} \Psi_{t}(\tau) \nabla_{\theta} \log \pi\left(a_{t} \mid s_{t} ; \theta_{k}\right),
\end{equation}
where $N$ denote the size of the trajectory set $\mathcal{T}_N$. Then the parameter $\theta$ is updated through $\theta_{k+1} = \theta_{k} + \eta \nabla_{\theta} \hat{J}(\pi_{\theta_{k}})$ where $\eta$ is the learning rate. For the Policy Newton method, the Hessian is similarly estimated as $\nabla_{\theta}^{2} \hat{J}(\theta_k) = \frac{1}{N} \sum_{\tau \in \mathcal{T}_{N}} H_{\theta}(\tau ; \pi_{\theta})$, and the policy is updated through $\theta_{k+1} = \theta_{k} + \eta[\nabla_{\theta}^{2} \hat{J}(\theta_k)]^{-1} \nabla_{\theta} \hat{J}(\theta_k)$, where $[\nabla_{\theta}^{2} \hat{J}(\theta_k)]^{-1} \nabla_{\theta} \hat{J}(\theta_k)$ is the Newton step. The calculation for the inverse of the Hessian is computationally unstable and costly. A direct way to alleviate this drawback is to introduce the regularization term and optimize an auxiliary function to obtain the Newton step \citep{Cubic_newton,regularization_Newton}:
\begin{equation}
\label{Auxiliary}
    \theta_{k+1}  =\underset{\theta \in \mathbb{R}^{d}}{\operatorname{argmin}}\left\{\left\langle\nabla_{\theta} \hat{J}\left(\theta_{k}\right), \theta-\theta_{k}\right\rangle \\
+\frac{1}{2}\left\langle\nabla_{\theta}^{2} \hat{J}\left(\theta_{k}\right)\left(\theta-\theta_{k}\right), \theta-\theta_{k}\right\rangle+\frac{\beta}{6}\left\|\theta-\theta_{k}\right\|^{3}\right\},
\end{equation}
where $\beta$ is the hyperparameter of the regularization term.
\subsection{Policy Gradient in RKHS}
Reproducing Kernel Hilbert Space (RKHS) is the vector valued Hilbert Space $\mathcal{H}_K$ where an elements $K(x,\cdot) \in \mathcal{H}_{K}$ satisfies the reproducing property $\langle K(x,\cdot), K(y,\cdot)\rangle = K(x,y)$. Despite the policy is modeled by the parameter $\theta$ with particular parameterized functions, the stochastic policy $\pi$ is directly modeled with a function $h$ in RKHS $\mathcal{H}_K$, where the updating gradient for it is also a function \citep{RKHS1}. Particularly, we denote the policy as $\pi_{h}\left(a_{t} \mid s_{t}\right) = \frac{1}{Z} e^{\mathcal{T} h(s_{t}, a_{t})}$ for discrete action space, where $Z = \sum_{a' \in \mathcal{A}} e^{\mathcal{T} f(s_t, a')}$ is the normalization constant and $\mathcal{T}$ is the temperature. Through the definition of the Fréchet derivative \citep{Frechet}, the Policy Gradient in RKHS is derived as \citep{RKHS_discrete,RKHS1, RKHS_TAC}:
\begin{equation}
\label{first_order_RKHS}
\begin{aligned}
    \nabla_{h} J(\pi_{h}) =& \mathbb{E}_{\tau \sim p(\tau ; \pi_{h})}\left[ \mathbb{g}_h(\tau;\pi_{h}) \right] =  \mathbb{E}_{\tau \sim p(\tau ; \pi_{h})}\left[ \sum_{t=0}^{T-1} \Psi_{t}(\tau) \nabla_{h} \log \pi_{h}\left(a_{t} \mid s_{t} \right)\right]\\
    =&\mathbb{E}_{\tau \sim p(\tau ; \pi_{h})}\left[ \sum_{t=0}^{T-1} \Psi_{t}(\tau) \mathcal{T}\left( K\left((s_{t}, a_{t}), \cdot\,\right) - \mathbb{E}_{a' \sim \pi_{h}(\cdot \mid s_{t})} \left[ K\left((s_{t}, a'), \cdot\,\right) \right] \right)\right],
\end{aligned}
\end{equation}
where $K\left((s_{t}, a_{t}), \cdot\,\right)$ is the kernel section induced by the state action pair $(s_{t},a_{t})$. Without loss of generality, the action space is set discretely in the rest of the paper. The estimation of the gradient is similar to Equation \eqref{Est_gradient}, where we denote it as $\nabla_{h} \hat{J}(\pi_{h}) = \frac{1}{N}\sum_{\tau \in \mathcal{T}_{N}}\mathbb{g}_h(\tau;\pi)$. Then the policy is updated iteratively by $h_{k+1} = h_{k} + \eta\nabla_{h} \hat{J}(\pi_{h})$. For simplicity, we denote $\hat{J}(\pi_{h})$ as $\hat{J}(h)$ in the rest of this paper.

The use of RKHS policy improves the sample efficiency greatly during training \citep{RKHS_TAC2}, and the overall performance is better than traditional gradient methods \citep{RKHS_NN}. However, there is still a lack of research on the Policy Newton algorithm within RKHS. A potential obstacle for its derivation is that the Hessian of $J(\pi_{h})$ is infinite, where its inverse is infeasible to represent explicitly. In previous research, Newton optimization is mainly studied in the online learning problem. In \citep{RKHS_Newton1}, the Newton optimization scheme is derived where the inverse of the Hessian in RKHS is approximated iteratively. The authors in \citep{RKHS_Newton2} integrate the adaptive embedding with the inverse approximation, which alleviates the computational burden. Despite this success in RKHS Newton optimization, the learning scheme is only suitable for data sampled from the same distribution during training \citep{RKHS_Newton3,RKHS_Newton4}, while the distribution of transitions in RL is related to the updating policy. As far as we investigated, this is the first paper to derive the Policy Newton in RKHS, and the convergence of our algorithm is also guaranteed.

\section{The Policy Newton in RKHS}
In this section, we present the derivation of the Policy Newton in RKHS. In the conventional Policy Newton method, it is simple to obtain the Hessian by deriving the second-order derivative of the expected discounted cumulative reward $J(\pi_{\theta})$ with respect to the parameter $\theta$. However, in RKHS space, the second-order Fréchet derivative may not be implicitly represented. Before the derivative of the Hessian within RKHS, we first introduce the following definition: 

\begin{definition}
\label{Define_RKHS}
    Defining the outer product $\mathcal{H}_{K} \otimes \mathcal{H}_{K}$  as a new RKHS with operator $\mathbb{K}((s_{t}, a_{t}),(s'_{t}, a'_{t})) = K\left((s_{t}, a_{t}), \cdot\,\right) \otimes K\left((s'_{t}, a'_{t}), \cdot\,\right) \in \mathcal{H}_{K} \otimes \mathcal{H}_{K}$ \citep{Tensor_RKHS}, it satisfies that: 
    \begin{equation}
    \nonumber
    \begin{aligned}
        \mathbb{K}((s_{t}, a_{t}),(s'_{t}, a'_{t})) \circ K\left((s''_{t}, a''_{t}), \cdot\,\right)& = K\left((s_{t}, a_{t}), \cdot\,\right) K\left((s'_{t}, a'_{t}), (s''_{t}, a''_{t})\right) \\
        <\mathbb{K}((s_{t}, a_{t}),(s'_{t}, a'_{t})), \mathbb{K}((s''_{t}, a''_{t}),(s'''_{t}, a'''_{t}))> &= K\left((s_{t}, a_{t}), (s''_{t}, a''_{t})\right) K\left((s'_{t}, a'_{t}), (s'''_{t}, a'''_{t})\right)
    \end{aligned}
    \end{equation}
\end{definition}
Intuitively, the Fréchet derivative $\nabla_{h} J(\pi_{h})$ is a mapping from $R \rightarrow \mathcal{H}_{K}$. Therefore, the second-order Fréchet derivative could naturally be a mapping from $\mathcal{H}_{K} \rightarrow \mathcal{H}_{K} \otimes \mathcal{H}_{K}$. We show in the following lemma that the Hessian within RKHS goes to the space of $\mathcal{H}_{K} \otimes \mathcal{H}_{K}$.
\begin{lemma}
\label{theorem_RKHS_second}
     The second-order Fréchet derivative $\nabla_{h}^2 J(\pi_{h}) =  \mathbb{E}_{\tau \sim p(\tau ; \pi_{h})}\left[ H_{h} (\tau;\pi_h) \right]$, where $H_{h} (\tau;\pi_h)$ is
\end{lemma}
\begin{equation}
\nonumber
   \left( \sum_{t=0}^{T-1} \Psi_t(\tau) \nabla_h \log \pi_h^t\right) \otimes \left( \sum_{t'=0}^{T-1} \nabla_h^\top \log \pi_h^t \right) 
    - \sum_{t=0}^{T-1} \Psi_t(\tau) \mathcal{T} \ \text{Cov}_{a' \sim \pi(\cdot \mid s_t)} \left[ K\left((s_{t}, a'_{t}), \cdot\,\right) \right].
\end{equation}
Here $\nabla_h \log \pi_h^t = \nabla_h \log \pi_h(a_t|s_t)$ and $\text{Cov}_{a' \sim \pi(\cdot \mid s_{t})} \left[ K\left((s_{t}, a'_{t}), \cdot\,\right) \right]$ denotes the covariance operator for kernel section $K\left((s_{t}, a'_{t}), \cdot\,\right)$, which is detailed as:
\begin{equation}
\nonumber
     \mathbb{E}_{a' \sim \pi(\cdot \mid s_{t})} \left[ K\left((s_{t}, a'), \cdot\,\right) \otimes K\left((s_{t}, a'), \cdot\,\right) \right] - \mathbb{E}_{a' \sim \pi(\cdot \mid s_{t})} K\left((s_{t}, a'), \cdot\,\right) \otimes \mathbb{E}_{a'' \sim \pi(\cdot \mid s_{t})} K\left((s_{t}, a''), \cdot\,\right).
\end{equation}
 The detailed derivation is shown in Appendix \ref{second_order detail}. Here, we only introduce a simple example,$U(h) =  e^{\mathcal{T} h(s_{t}, a_{t})} K\left((s_{t}, a_{t}), \cdot\,\right)$, which is a component in  $\nabla_{h} J(\pi_{h})$, to present the core concept for introducing the outer product in RKHS when implementing the Fréchet derivative.

 \paragraph{The second-order Fréchet derivative} Let $h,g \in \mathcal{H}_{K}$ and $\mathbb{D}(h) =  \mathcal{T}e^{\mathcal{T} h(s_{t}, a_{t})} K\left((s_{t}, a_{t}), \cdot\,\right) \otimes K\left((s_{t}, a_{t}), \cdot\,\right)$. Then according to the definition of the Fréchet derivative \citep{Frechet}, we testify that $D(h) = \nabla_h U(h)$: 
 \begin{equation}
 \nonumber
 \begin{aligned}
     \frac{||U(h+g) - U(h)-\mathbb{D}(h) \circ g||}{||g||} =  \frac{||e^{\mathcal{T} h(s_{t}, a_{t})} K\left((s_{t}, a_{t}), \cdot\,\right)\left[e^{\mathcal{T} g(s_{t}, a_{t})}-1 - g(s_{t}, a_{t})\right]||}{||g||} \\
     = \frac{||e^{\mathcal{T} h(s_{t}, a_{t})} K\left((s_{t}, a_{t}), \cdot\,\right)\frac{Cg^2(s_{t}, a_{t})}{2}||}{||g||} \leq K\left((s_{t}, a_{t}), (s_{t}, a_{t})\right)||e^{\mathcal{T} h(s_{t}, a_{t})} \frac{Cg(s_{t}, a_{t})}{2}|| \xrightarrow{g \rightarrow 0} 0 ,
 \end{aligned}
 \end{equation}
where the last inequality is due to  Cauchy-Schwarz. Through Lemma \ref{theorem_RKHS_second}, we could find that the second-order Fréchet derivative $\nabla_{h}^2 J(\pi_{h})$ is infeasible to present explicitly. Calculating its inverse is further infeasible for the Policy Newton methods in RKHS. However, we can still obtain the RKHS Newton step $\Delta h$ through optimizing the regularized auxiliary function similar to Equation \eqref{Auxiliary}:
\begin{equation}
\label{RKHS_Auxiliary}
        \Delta h  =\underset{\bar{h} \in \mathcal{H}_{K}}{\operatorname{argmin}}\left\{\left\langle\nabla_{h} \hat{J}\left(h_{k}\right), \bar{h}\right\rangle \\
+\frac{1}{2}\left\langle\nabla_{h}^{2} \hat{J} \left(h_{k}\right)\circ \bar{h}, \bar{h}\right\rangle+\frac{\beta}{6}\left\|\bar{h}\right\|^{3}\right\},
\end{equation}
where $\nabla_{h}^{2} \hat{J}\left(h_{k}\right) =\frac{1}{N} \sum_{\tau \in \mathcal{T}_{N}} H_{h} (\tau;\pi_h)$ is the estimated RKHS Hessian. Although this estimation is not computationally feasible, the corresponding second-order component $\left\langle\nabla_{h}^{2} \hat{J} \left(h_{k}\right)\circ\bar{h}, \bar{h}\right\rangle$ is easy to calculate according to the Definition \ref{Define_RKHS}. The optimization in $\mathcal{H}_{K}$ is still hard to proceed, but through the Representer Theorem in RKHS, we could easily transform the parameter space in this optimization problem from $\mathcal{H}_{K}$ into $\mathbb{R}$.
\begin{lemma}
\label{Representer}
    \textbf{Representer Theorem} \citep{RepresenterTheorem}. Suppose we are given a nonempty set $\mathcal{X}$, a positive definite real-valued kernel $K(\cdot,\cdot)$ on $\mathcal{X} \times \mathcal{X}, a$ training sample $\left(x_1, y_1\right), \ldots,\left(x_M, y_M\right) \in \mathcal{X} \times \mathbf{R}$, a strictly monotonically increasing real-valued function $\mathcal{G}$, an arbitrary cost function $c$.  Then any $h \in \mathcal{H}_{K}$ minimizing the regularized functional
\begin{equation}
\nonumber
    c\left(\left(x_1, y_1, h\left(x_1\right)\right), \ldots,\left(x_M, y_M, h\left(x_M\right)\right)\right)+\mathcal{G}(\|h\|)
\end{equation}

admits a representation of the form $h(\cdot)=\sum_{i=1}^M \alpha_i K\left(x_i, \cdot \right)$, 
where $\alpha_i$ is the weight for kernel sections.
\end{lemma}
Applying the Representer Theorem (Lemma~\ref{Representer}), Optimization Problem~\ref{RKHS_Auxiliary} is equivalent to finding $\boldsymbol{\alpha}^{*}$ via:
\begin{equation}
\label{transformed_RKHS}
\boldsymbol{\alpha}^{*} =\underset{\boldsymbol{\alpha} \in \mathbb{R}^{NT}}{\operatorname{argmin}}\left\{\left\langle\nabla_{h} \hat{J}\left(h_{k}\right), \bar{h}_{\boldsymbol{\alpha}}\right\rangle  +\frac{1}{2}\left\langle\nabla_{h}^{2} \hat{J} \left(h_{k}\right)\circ \bar{h}_{\boldsymbol{\alpha}}, \bar{h}_{\boldsymbol{\alpha}}\right\rangle+\frac{\beta}{6}\left\|\bar{h}_{\boldsymbol{\alpha}}\right\|^{3}\right\},
\end{equation}
where $\bar{h}_{\boldsymbol{\alpha}} = \sum_{i=1}^{N }\sum_{t=1}^{T} \alpha_t^i K\left((s_t^i,a_t^i), \cdot \right)$. Here, $(s_t^i,a_t^i)$ denotes the state-action pair for the $i$-th trajectory at time step $t$, and $\boldsymbol{\alpha} = \{\alpha_t^i\}_{i=1, t=1}^{N, T}$ is the set of kernel weights. These weights can be vectorized as $\boldsymbol{\bar{\alpha}} \in \mathbb{R}^{NT}$, where the $l$-th element corresponds to $\alpha_t^i$ with $k = (i-1)T + t$. Along with $\boldsymbol{\bar{\alpha}}$, this Optimization Problem~\eqref{transformed_RKHS} can be simplified to a standard form, stated in the following theorem.

\begin{theorem}
\label{theorem_final}
    The optimization of the Policy Newton step in RKHS is equal to the optimization of the following quadratic optimization with cubic regularization:
    \begin{equation}
    \label{optimization_final}
    \boldsymbol{\bar{\alpha}}^{*} =\underset{\boldsymbol{\bar{\alpha}} \in \mathbb{R}^{NT}}{\operatorname{argmin}}\left\{\left\langle v, \boldsymbol{\bar{\alpha}}\right\rangle  +\frac{1}{2}\left\langle H \boldsymbol{\bar{\alpha}}, \boldsymbol{\bar{\alpha}}\right\rangle+\frac{\beta}{6}\left\|\boldsymbol{\bar{\alpha}}\right\|_2^{3}\right\}.
\end{equation}
Here, $v \in \mathbb{R}^{NT}$ is the first-order coefficient vector where 
\begin{equation}
\nonumber
    v_i = \frac{\mathcal{T}}{N} \sum_{l=1}^{NT} \Psi_l(\tau) \left( K\left( (s_l, a_l), (s_i, a_i) \right) - \mathbb{E}_{a'} \left[ K\left( (s_l, a'), (s_i, a_i) \right) \right] \right).
\end{equation}
Let $H \in \mathbb{R}^{NT \times NT}$ be the second-order coefficient matrix given by:
\begin{equation}
\label{eq:H_definition} 
  H = \frac{\mathcal{T}^2}{N} b c^\top - \frac{\mathcal{T}}{N}\sum_{l=1}^{NT} \Psi_l(\tau)  \Sigma^{(l)}.
\end{equation}
Here, $b \in \mathbb{R}^{NT}$ and $c \in \mathbb{R}^{NT}$ are vectors, and $\Sigma^{(l)}$ represents component-related covariance information. The components $b_i$, $c_i$, and the related covariance terms $\Sigma^{(l)}_{ij}$ are defined as:
\begin{equation}
\nonumber
    \left\{
\begin{aligned} 
  b_i &= \sum_{l=1}^{NT} \Psi_t(\tau) \left( K_{it} - \mathbb{E}_{a'} [ K_{it}' ] \right ), && \quad c_i = \sum_{l=1}^{NT} \left( K_{il} - \mathbb{E}_{a'} [ K_{il}' ] \right ) \\
  \Sigma^{(l)}_{ij} &= \mathrm{Cov}_{a' \sim \pi(\cdot \mid s_l)} \left[ K_{il}', K_{jl}' \right ] && \\
\end{aligned}
\right.
\end{equation}
In these expressions, the kernel terms are $K_{il} = K\left( (s_i, a_i), (s_l, a_l) \right)$ and $K_{il}' = K\left( (s_i, a_i), (s_l, a') \right)$.
\end{theorem}
The detailed derivation for Theorem \ref{theorem_final} is presented in Appendix \ref{Derivation_theorem_final}. It is observed from this theorem that the Policy Newton in RKHS is similar to the traditional Policy Newton method, but the complexity of the Optimization Problem \eqref{optimization_final} is dependent on the volume of data, i.e., $N \times T$, which possesses the same property of other RKHS methods like support vector machine \citep{SVM} and radial basis function networks \citep{RBFN}. We show in the next section the suboptimality and convergence rate for the proposed Policy Newton in RKHS. 


\section{The suboptimality and convergence rate of Policy Newton in RKHS}
\label{Theorem_sec}
In this section, we first detail the Policy Newton in RKHS algorithm. We then analyze its convergence properties, demonstrating that despite optimizing via a surrogate function, the resulting policy converges to a local optimum. Furthermore, we show that our proposed Policy Newton in RKHS exhibits a second-order convergence rate, in contrast to the first-order rate achieved by Policy Gradient in RKHS.

\subsection{The Policy Newton in RKHS method}
The Optimization Problem~\eqref{transformed_RKHS} admits two primary solution approaches:

(1) Directly computing the derivative of the objective function, setting it to zero, and solving for the critical points.

(2) Optimizing it using various classic optimization methods, including gradient descent, the Newton method, and the conjugate gradient method~\citep{conjugate}.

While the analytic method (1) is conceptually simple and direct, it can introduce significant instability into the training process. In complex environments, this instability can lead to exponential error growth. Consequently, method (2) represents a more practical optimization approach. We select the conjugate gradient method as our optimization method. Its specific settings are detailed in Appendix \ref{optimization_detail}.

\begin{algorithm}
\caption{Policy Newton RKHS Method}
\label{alg:RKHS_Newton} 
\textbf{Input:} Number of iterations $M$, trajectory batch size $N$, learning rate $\eta$
\begin{algorithmic}[1] 
    \STATE Initialize RKHS function $h_1 \leftarrow 0$, actor policy $\pi_{h_1}$ based on $h_1$, trajectory set $\mathcal{T}$.
    \FOR{$m = 1, \dots ,M$}
        \STATE Sample $N$ trajectories using the current policy $\pi_{h_m}$, store in $\mathcal{T}$.
        \STATE Estimate the first-order coefficient vector $v$ and second-order coefficient matrix $H$ using $\mathcal{T}$ (according to Theorem \ref{theorem_final}).
        \STATE Solve the Optimization Problem~\eqref{optimization_final} using conjugate gradient descent method, output the optimization result $\boldsymbol{\bar{\alpha}}$.
        \STATE Construct the RKHS update step $\Delta h$ using $\boldsymbol{\bar{\alpha}}$ via Lemma~\ref{Representer}.
        \STATE Update the RKHS function: $h_{m+1} \leftarrow h_m + \eta\Delta h$ and the actor policy $\pi_{h_{m+1}}$ based on $h_{m+1}$.
    \ENDFOR
    \RETURN final policy $\pi_{h_{M+1}}$. 
\end{algorithmic}
\end{algorithm}
\subsection{Suboptimality analysis of the proposed algorithm}
Theoretically, we assume that the optimal solution to Problem~\eqref{optimization_final} is consistently achieved by Algorithm~\ref{alg:RKHS_Newton}. To establish the convergence properties of our algorithm, we introduce the following lemmas and assumptions.

\begin{lemma}[Monte Carlo convergence rate (\ref{MC_convergence})]
     Assuming $\mathbb{E}_{\tau \sim p(\tau ; \pi_h)}[||\mathbb{g}_h(\tau;\pi_h)||^2]\leq \sigma^2_{0}$ and $\mathbb{E}_{\tau \sim p(\tau ; \pi_h)}[||H_{h}(\tau;\pi_h)||^2]\leq \sigma^2_{1}$, the Monte Carlo estimation of first and second-order Fréchet derivative, namely $\nabla_{h} \hat{J}\left(h_{k}\right)$ and $\nabla_{h}^2 \hat{J}\left(h_{k}\right)$, achieve the convergence rate of $O(\frac{1}{\sqrt{N}})$:
    \begin{equation}
    \nonumber
        \mathbb{E}_{\tau \sim p(\tau ; \pi_h)} \left[||\nabla_{h} \hat{J}\left(h_{k}\right)-\nabla_{h} J\left(h_{k}\right)||^2 \right] \leq \frac{\sigma^2_{0}}{N} , \; \mathbb{E}_{\tau \sim p(\tau ; \pi_h)} \left[||\nabla_{h}^2 \hat{J}\left(h_{k}\right)-\nabla_{h}^2 J\left(h_{k}\right)||^2 \right] \leq \frac{\sigma^2_{1}}{N}.
    \end{equation}  
\end{lemma}
The proof is straightforward by using the property of expectation, which we show in the Appendix \ref{MC_convergence}. Establishing convergence also requires a Lipschitz continuity assumption for the Hessian, similar to other gradient-based methods \citep{policy_gradient1, policy_gradient2}.

\begin{assumption}[Lipschitz continuous]
\label{Lipschitz}
     The Hessian operator $\nabla_h^2 J(h)$ is Lipschitz continuous with constant $0\leq L \leq \beta$, i.e., for all $h_1, h_2 \in \mathcal{H}_K$:
\begin{equation}\nonumber
\| \nabla_h^2 J(h_1) - \nabla_h^2 J(h_2) \| \le L \|h_1 - h_2\|.
\end{equation}
\end{assumption}
Through this assumption, we could establish the upper bound for $J(h)$ with respect to the Hessian and step norm:
\begin{lemma}[Taylor upper bound (\ref{Taylor upper bound})]
\label{Taylor_uppper}
     Under Assumption \ref{Lipschitz}, for any $h_1, h_2 \in \mathcal{H}_K$:
\begin{equation}\nonumber
J(h_2) \le J(h_1) + \langle \nabla_h J(h_1), h_2 - h_1 \rangle + \frac{1}{2} \langle \nabla_h^2 J(h_1) \circ (h_2 - h_1), h_2 - h_1 \rangle + \frac{L}{6} \|h_2 - h_1\|^3.
\end{equation}
\end{lemma}
The fundamental approach to proving convergence centers on establishing a relationship between the expected gradient norm, $\mathbb{E}\| \nabla_h J(h_{k})\|$, and a function denoted by $L(\beta, \sigma^2_0, \sigma^2_1, N)$. Following standard techniques in convergence analysis, the norm of the update step, $\| h_{k+1} - h_k \|$, is employed as an intermediate quantity to construct this relationship. To this end, an upper bound for this step norm is derived in Lemma \ref{upper bound}.
\begin{lemma}[Step norm upper bound (\ref{proof_step})]
    \label{upper bound}
     Denoting the updating times for Policy Newton in RKHS as $M$, and the number of trajectories sampled in each updating as $N$, the updating step can be upper bounded as:
    \begin{equation}\nonumber
        \mathbb{E}\left[\|h_{R+1}-h_R\|^3\right] \leq \frac{36(J(h_1) - J^*)}{\beta M} + \frac{48\sqrt{3}}{\beta^{3/2}} \frac{\sigma_0^{3/2}}{N^{3/4}} + \frac{864}{\beta^3} \frac{\sigma_1^3}{N^{3/2}},
    \end{equation}
    where $R$ is a random variable uniformly distributed on $\{1, \ldots, M\}$, such that $P(R=k) = 1/M$.
\end{lemma}
This lemma provides an upper bound on the expected cubed norm of the update step involving a randomly selected iteration $R$, which is a key quantity used subsequently to establish convergence bounds in expectation for the Policy Newton in RKHS. Next, we establish the lower bound of the norm for the update step, which relates it to the RKHS gradient.
\begin{lemma}[Step norm lower bound (\ref{proof_step_lower})]
    \label{lower bound}
     The updating step can be lower bounded as:
    \begin{equation}\nonumber
         \mathbb{E}[\|h_{k+1}-h_k\|^2] \geq \frac{1}{L + \beta} \left( \mathbb{E}[\|\nabla J(h_{k+1})\|] - \frac{\sigma_0}{\sqrt{N}} - \frac{\sigma_1^2}{2N(L+\beta)} \right).
    \end{equation}
\end{lemma}
Having established both lower and upper bounds for the step norm, we can now use these results to construct the main convergence theorem. 
\begin{theorem}[Convergence property (\ref{Convergence})]         Given Lemmas \ref{upper bound} and \ref{lower bound}, let $R$ be a random variable uniformly distributed on $\{1, \ldots, M\}$. The sequence $\{h_k\}$ generated by iterative optimization in Theorem \ref{theorem_final} satisfies
    \begin{equation*}
        \lim_{M, N \to \infty} \mathbb{E}[\|\nabla J(h_{R+1})\|] = 0.
    \end{equation*}
\end{theorem}
This theorem indicates that the expected gradient norm at a randomly chosen iteration converges to zero, implying convergence towards a stationary point. 
\subsection{The second-order convergence rate}

Establishing the convergence rate for stochastic Policy Newton methods typically requires specific assumptions regarding problem structure, including strong convexity, Lipschitz smoothness, carefully designed step-size schedules, and bounded variance of gradient estimators \citep{SIAM_Opt1, SIAM_Opt2}. The convergence behaviors range from sublinear to linear or even superlinear under various stochastic settings and assumptions. Therefore, to establish baseline performance characteristics, this section analyzes the convergence rate under idealized conditions, while a comprehensive analysis under more realistic stochastic assumptions remains for future work. Specifically, we assume access to the true gradient $\nabla_{h} J(h_k)$ and Hessian $\nabla_h^2 J(h_k)$ at each iteration $k$, effectively setting $\nabla_{h} \hat{J}(h_k) = \nabla_{h} J(h_k)$ and $\nabla_{h}^2 \hat{J}(h_k) = \nabla_{h}^2 J(h_k)$.
Under this deterministic scenario, we prove that the method achieves a local quadratic convergence rate.

\begin{theorem}[Local Quadratic Convergence (\ref{convergence_rate_proof})]
 Consider the deterministic Policy Newton RKHS method (Algorithm \ref{alg:RKHS_Newton} with $\nabla_{h} \hat{J}(h_k) = \nabla_{h} J(h_k)$ and $\nabla_{h}^2 \hat{J}(h_k) = \nabla_{h}^2 J(h_k)$). 
 Assuming the norm of the inverse operator $\|(\nabla_{h}^2 J(h_k) + \frac{\beta}{2} \|\Delta h_k\| \mathcal{I})^{-1}\|$ is bounded by some constant $B$, and we assume that the update step is sufficiently small that $\|\Delta h_k\| \le K \|e_k\|$ for some $K > 0$.
 
 If the initial iterate $h_0$ is sufficiently close to $h^*$, the sequence $\{h_k\}$ converges quadratically to $h^*$. That is, there exists a constant $C > 0$ such that
 $$ \|h_{k+1} - h^*\| \leq C_q \|h_k - h^*\|^2 $$
 for all $k$ sufficiently large.
\end{theorem}

\section{Numerical experiment}
This section presents an empirical evaluation of our proposed Policy Newton method in Reproducing Kernel Hilbert Space (RKHS) across two distinct experimental settings: (a) a simplified Asset Allocation environment designed specifically to demonstrate the quadratic convergence properties of Policy Newton in RKHS \citep{Asset_Allocation2}, and (b) complex control tasks from the Gymnasium framework \citep{Gymnasium}, including CartPole and Lunar Lander. The Asset Allocation environment serves to empirically validate the theoretical convergence guarantees established in Section \ref{Theorem_sec}. Additionally, we benchmark Policy Newton in RKHS against several baseline methods in complex environments to demonstrate its superior performance characteristics.

\subsection{Quadratic convergence tested in the toy experiment}
We empirically validate the quadratic convergence properties of Policy Newton in RKHS using a simplified Asset Allocation environment \citep{Asset_Allocation1, Asset_Allocation2}. While the complete asset allocation problem presents substantial analytical challenges, we utilize a reduced-complexity variant (detailed in Appendix \ref{asset_allocation}) where the global optimal policy can be explicitly represented, enabling precise quantification of convergence properties for both the policy and cumulative reward $J(\pi)$.

\begin{figure}[htbp]
    \centering
    \begin{subfigure}[b]{0.49\linewidth}
        \centering
        \includegraphics[width=\columnwidth]{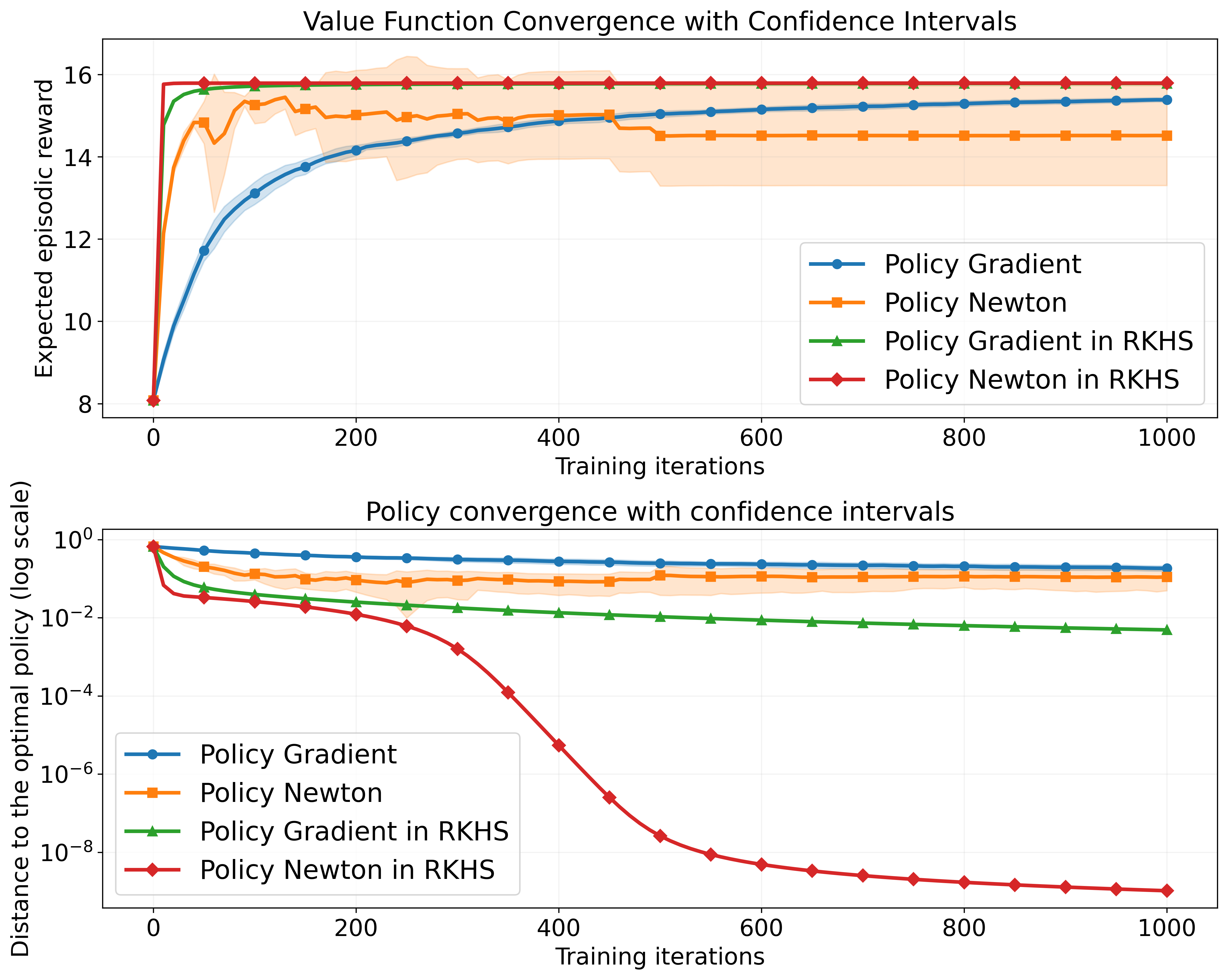}
        \caption{  }
        \label{convergence_train}
    \end{subfigure}
    \hfill
    \begin{subfigure}[b]{0.49\linewidth}
        \centering
        \includegraphics[width=\columnwidth]{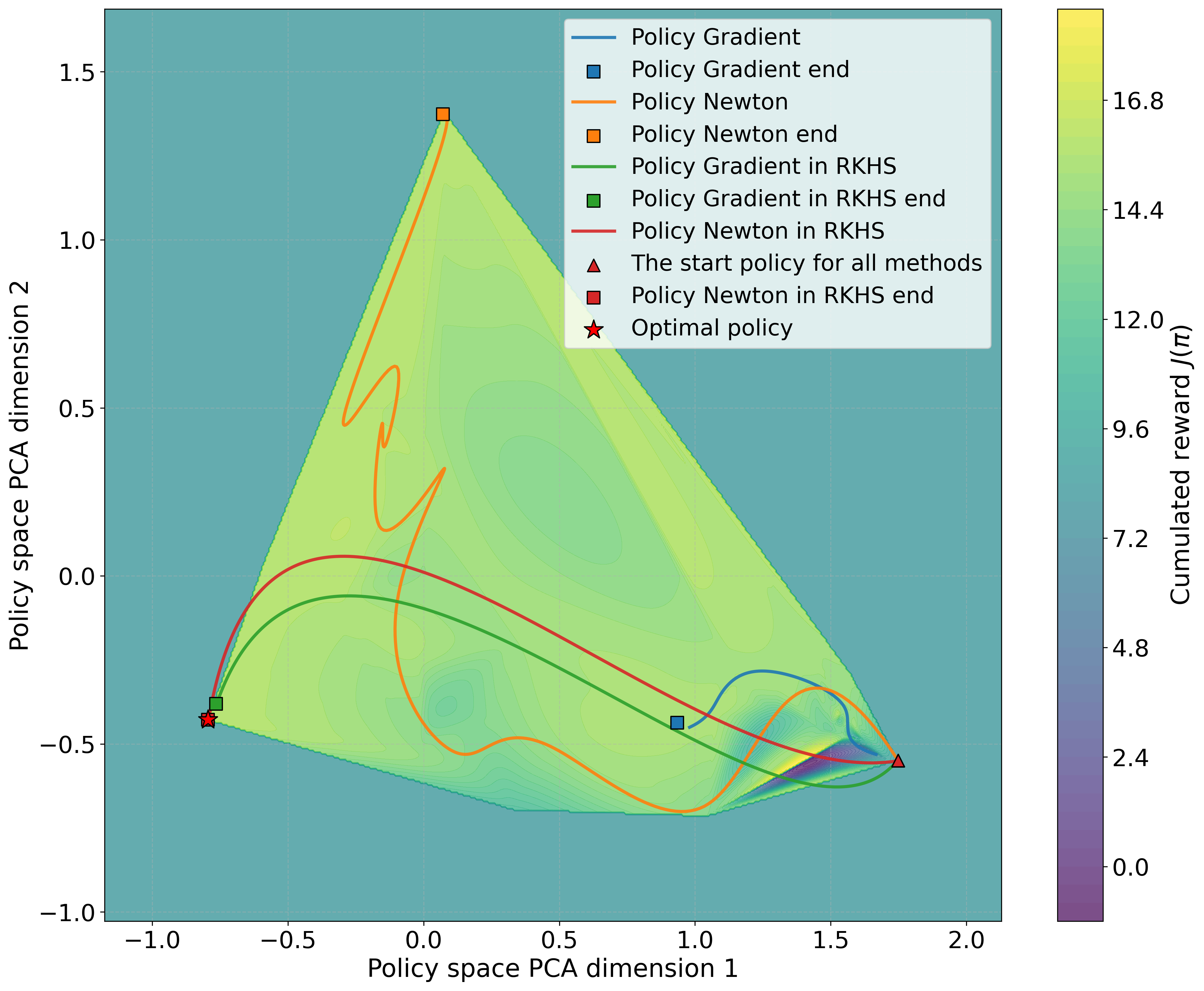}
        \caption{  }
        \label{trajectory}
    \end{subfigure}
    \caption{Experimental results demonstrating quadratic convergence of Policy Newton in RKHS within the simplified Asset Allocation environment. Three benchmark methods are compared: conventional Policy Gradient, Policy Newton, and Policy Gradient in RKHS. \ref{convergence_train} illustrates the convergence metrics during training, while \ref{trajectory} shows the optimization trajectories on the reward surface.}\label{visualize_gradient}
\end{figure}

For comparative analysis, we implemented four distinct methodologies. The conventional Policy Gradient and Policy Newton methods \citep{Cubic_newton} utilize discrete policies with parameterized action probabilities. The Policy Gradient in RKHS implementation follows the approach described in \citep{RKHS1}, while our Policy Newton in RKHS method is implemented according to Algorithm \ref{alg:RKHS_Newton}. All policies were initialized with uniform distributions.

The experimental results presented in Figure \ref{visualize_gradient} reveal several important findings. Figure \ref{convergence_train} demonstrates that both Policy Gradient in RKHS and Policy Newton in RKHS converge rapidly toward the maximum expected episodic reward. Notably, Policy Newton in RKHS exhibits clear quadratic convergence behavior as it approaches the optimal policy. In contrast, the conventional methods show different characteristics—while the standard Policy Newton method converges more rapidly than conventional Policy Gradient (as shown in Figures \ref{convergence_train} and \ref{trajectory}), its training trajectory exhibits greater instability and ultimately converges to a suboptimal local maximum. These results highlight how policy optimization in RKHS effectively leverages infinite-dimensional feature representations, enabling the optimization process to escape local optima that constrain conventional methods.

\subsection{Training performance in RL testing environment}
\begin{figure}[htbp]
    \centering
    \begin{subfigure}[b]{0.49\linewidth}
        \centering
        \includegraphics[width=\columnwidth]{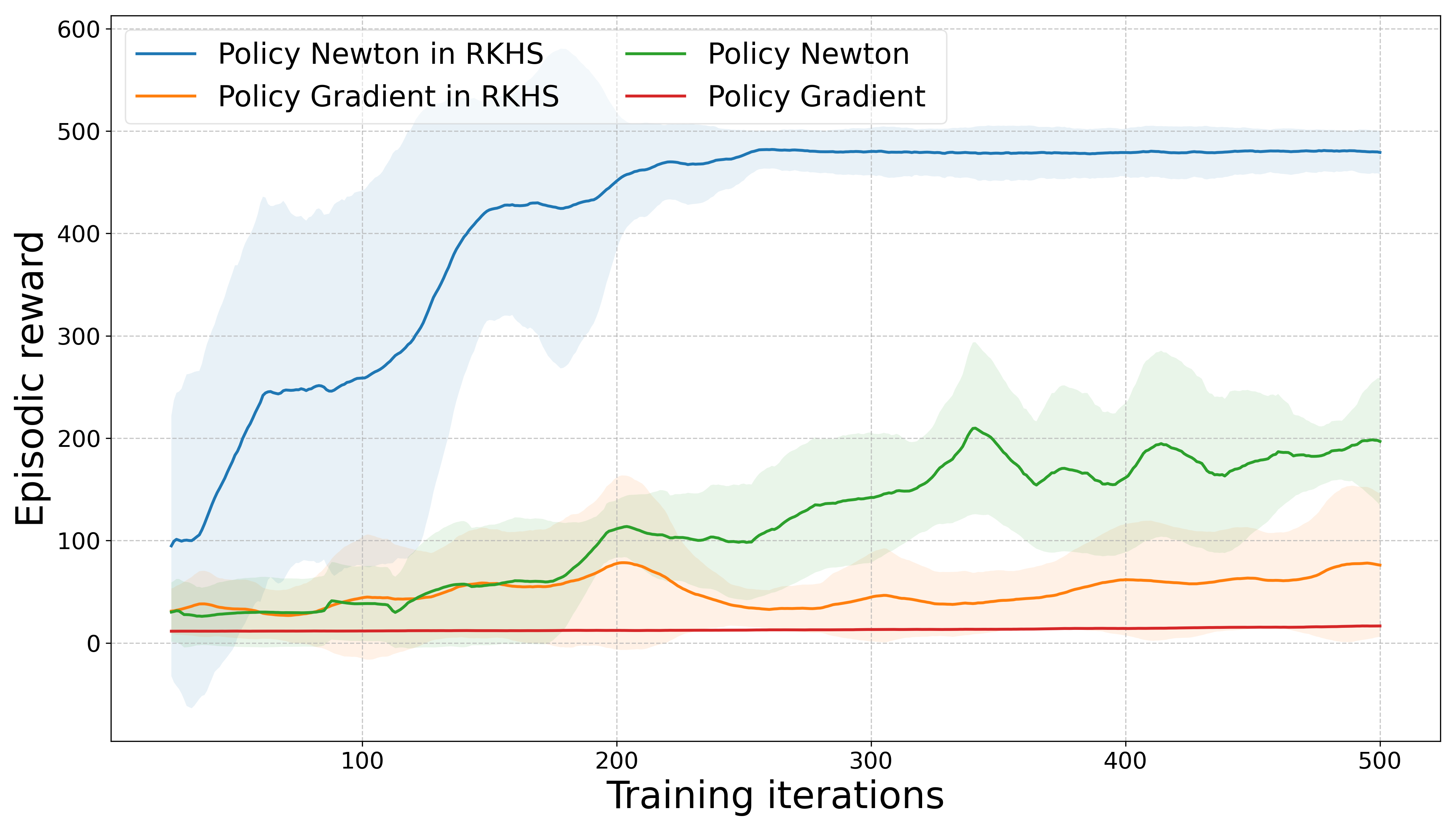}
        \caption{CartPole}
        \label{CartPole_plot}
    \end{subfigure}
    \hfill
    \begin{subfigure}[b]{0.49\linewidth}
        \centering
        \includegraphics[width=\columnwidth]{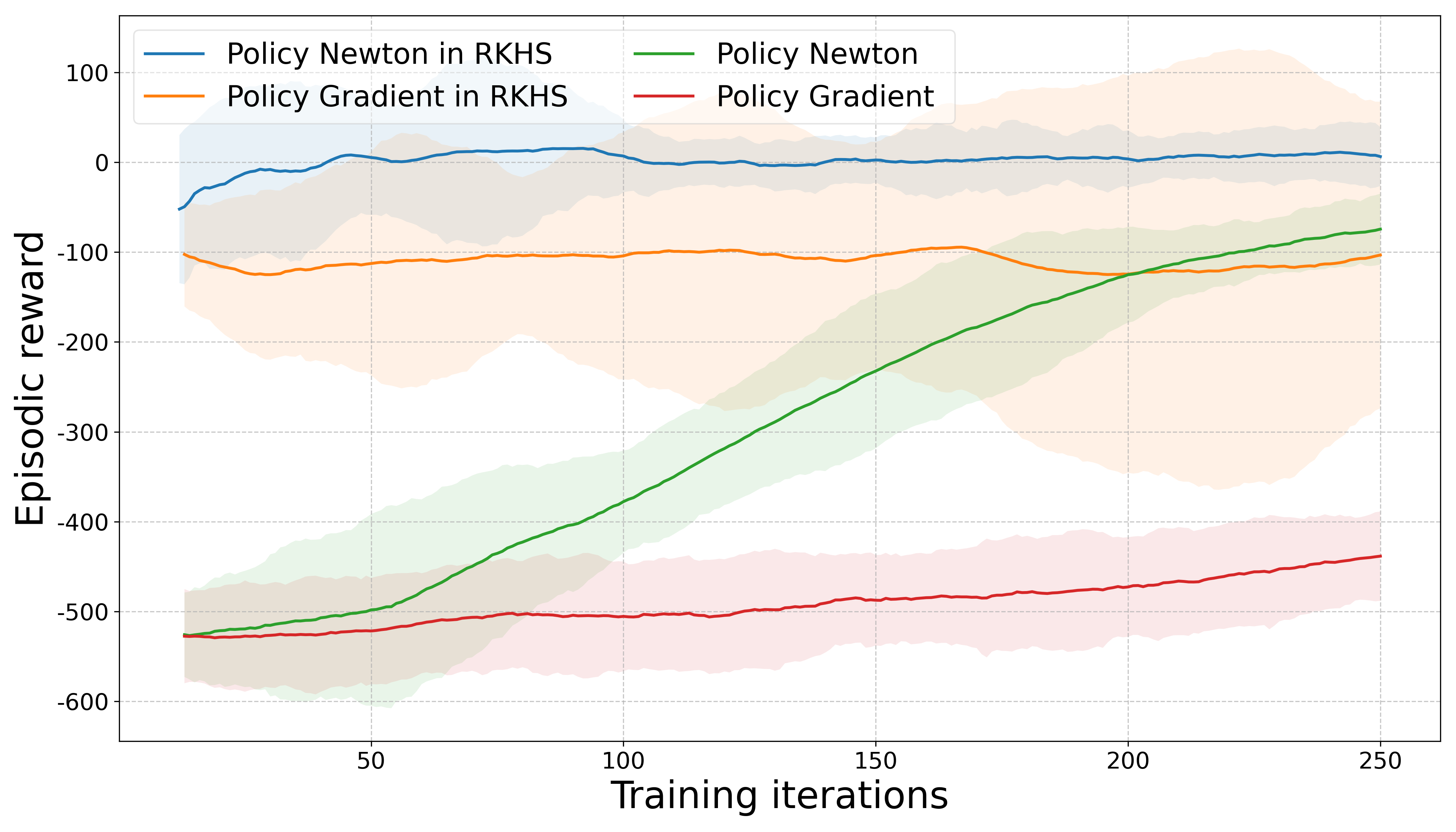}
        \caption{Lunar Lander}
        \label{Lunar Lander_plot}
    \end{subfigure}
    \caption{Comparative analysis of Policy Newton in RKHS against established baseline methods in Gymnasium environments. The plots display mean episodic rewards throughout the training process, demonstrating the superior sample efficiency and asymptotic performance of our proposed method. (a) Results from the CartPole balancing task. (b) Results from the more challenging Lunar Lander environment.}\label{Performance_final}
\end{figure}

To evaluate the efficacy of the proposed Policy Newton in RKHS algorithm, we conduct experiments utilizing two standard RL environments from the Gymnasium suite \citep{Gymnasium}: CartPole and Lunar Lander. These benchmarks serve to demonstrate the training performance of our method. Given that the observation spaces within these environments are continuous, the policies for both Policy Gradient and Policy Newton methods are parameterized using a linear model augmented with a polynomial transformation. This specific parameterization aligns with the implementation detailed in \citep{Cubic_newton}, wherein the polynomial transformation facilitates richer feature representations. More details are described in Appendix \ref{optimization_detail}.

As illustrated in Figure \ref{Performance_final}, the Policy Newton in RKHS method exhibits rapid convergence to the highest episodic reward. In contrast, the standard Policy Newton method achieves only comparable or inferior performance, requiring significantly more training iterations. Furthermore, both the Policy Gradient in RKHS and the standard Policy Gradient methods demonstrate limited convergence, suggesting that a substantially greater number of training steps would be necessary to reach the optimal reward.

\section{Conclusion and future work}
\label{sec:conclusion}
This paper successfully introduced Policy Newton in RKHS, the first practical second-order optimization method for reinforcement learning policies represented within Reproducing Kernel Hilbert Spaces. We established its theoretical foundations, proving convergence to a local optimum and demonstrating a local quadratic convergence rate. These theoretical properties were empirically validated on both a toy problem and standard RL benchmarks, where Policy Newton in RKHS achieved significantly faster convergence to superior episodic rewards compared to first-order and parametric Newton baselines.

While the current results are promising, extending the application of Policy Newton in RKHS to highly complex RL problems, such as the Humanoid environment \citep{Mujoco}, may reveal challenges related to robustness and stability. A promising avenue for future research is the integration of neural networks with the Policy Newton in RKHS framework, potentially drawing inspiration from architectures similar to those proposed in \citep{RKHS_NN}, to enhance performance in such demanding scenarios. The primary focus of this paper has been the rigorous theoretical establishment of Policy Newton in RKHS, laying the groundwork for these and other exciting explorations in future work.

\bibliographystyle{plainnat}  
\bibliography{references.bib}


\appendix


\section{The derivation of the second-order Fréchet derivative in RKHS}
\label{second_order detail}

This section details the derivation of second-order Fréchet derivatives of the log-policy, $\log \pi_h(a_t \mid s_t)$, with respect to the function $h \in \mathcal{H}_K$. The first-order Fréchet derivative is introduced in \cite{RKHS_discrete}, for convenience's sake, we also detail it in this section.  The policy is defined as $\pi_{h}\left(a_{t} \mid s_{t}\right) = \frac{1}{Z} e^{\mathcal{T} h(s_{t}, a_{t})}$, where $Z = \sum_{a' \in \mathcal{A}} e^{\mathcal{T} h(s_t, a')}$ is the normalization constant. For brevity in this section, we will denote $(s_t, a_t)$ as $(s,a)$ when the context is clear for a single state-action pair for which the log-policy is being differentiated. The kernel section $K((s,a), \cdot)$ is an element in $\mathcal{H}_K$.

We first derive the first-order Fréchet derivative $\nabla_h \log \pi_h(a \mid s)$.
The log-policy is $\log \pi_h(a \mid s) = \mathcal{T} h(s,a) - \log Z$.

The Fréchet derivative of the first term is:
\begin{equation}
\nonumber
\nabla_h (\mathcal{T} h(s,a)) = \mathcal{T} K((s,a), \cdot)
\end{equation}

For the second term, $-\log Z$:
\begin{equation}
\nonumber
\nabla_h (-\log Z) = -\frac{1}{Z} \nabla_h Z
\end{equation}
We compute $\nabla_h Z$:
$$Z = \sum_{a' \in \mathcal{A}} e^{\mathcal{T} h(s, a')}$$
\begin{equation}
\nonumber
\nabla_h Z = \sum_{a'} \nabla_h (e^{\mathcal{T} h(s, a')}) = \sum_{a'} e^{\mathcal{T} h(s, a')} \mathcal{T} \nabla_h h(s, a') = \mathcal{T} \sum_{a'} e^{\mathcal{T} h(s, a')} K((s, a'), \cdot)
\end{equation}
Substituting this back:
\begin{equation}
\nonumber
\nabla_h (-\log Z) = -\frac{1}{Z} \mathcal{T} \sum_{a'} e^{\mathcal{T} h(s, a')} K((s, a'), \cdot)
\end{equation}
Recognizing that $\frac{e^{\mathcal{T} h(s, a')}}{Z} = \pi_h(a' \mid s)$, we simplify:
\begin{equation}
\nonumber
\nabla_h (-\log Z) = -\mathcal{T} \sum_{a'} \pi_h(a' \mid s) K((s, a'), \cdot) = -\mathcal{T} \mathbb{E}_{a' \sim \pi_h(\cdot \mid s)} \left[ K((s, a'), \cdot) \right]
\end{equation}
Combining the derivatives of both terms, we obtain the first-order Fréchet derivative of the log-policy:
\begin{equation}
\nonumber
\nabla_h \log \pi_h(a \mid s) = \mathcal{T} K((s, a), \cdot) - \mathcal{T} \mathbb{E}_{a' \sim \pi_h(\cdot \mid s)} \left[ K((s, a'), \cdot) \right]
\end{equation}
Factorizing $\mathcal{T}$:
\begin{equation} \label{eq:appendix_first_order_log_policy}
\nabla_h \log \pi_h(a \mid s) = \mathcal{T} \left( K((s, a), \cdot) - \mathbb{E}_{a' \sim \pi_h(\cdot \mid s)} \left[ K((s, a'), \cdot) \right] \right)
\end{equation}
This expression forms the core of the Policy Gradient in RKHS as shown in Equation~\eqref{first_order_RKHS} of the main paper when appropriately weighted and summed.

Next, we derive the second-order Fréchet derivative (Hessian operator) of the log-policy with respect to $h$, denoted as $\nabla_h^2 \log \pi_h(a \mid s)$. This is obtained by differentiating Equation~\eqref{eq:appendix_first_order_log_policy}:
\begin{equation}
\nonumber
\nabla_h^2 \log \pi_h(a \mid s) = \nabla_h \left[ \mathcal{T} \left( K((s, a), \cdot) - \sum_{a' \in \mathcal{A}} \pi_h(a' \mid s) K((s, a'), \cdot) \right) \right]
\end{equation}
Since $\mathcal{T}$ is a constant and $K((s,a),\cdot)$ is a fixed element in $\mathcal{H}_K$ (not depending on $h$ for this differentiation), its derivative is zero:
\begin{equation}
\nonumber
\nabla_h^2 \log \pi_h(a \mid s) = \mathcal{T} \nabla_h \left( - \sum_{a' \in \mathcal{A}} \pi_h(a' \mid s) K((s, a'), \cdot) \right)
\end{equation}
Applying the product rule for Fréchet derivatives (treating $K((s,a'),\cdot)$ as a constant vector in $\mathcal{H}_K$ for each $a'$):
\begin{equation}
\nonumber
\nabla_h^2 \log \pi_h(a \mid s) = -\mathcal{T} \sum_{a' \in \mathcal{A}} \left( \nabla_h \pi_h(a' \mid s) \right) \otimes K((s, a'), \cdot)
\end{equation}
Here, $\otimes$ denotes the outer product as defined in Definition~\ref{Define_RKHS}.

Now, we compute $\nabla_h \pi_h(a' \mid s)$. Recall $\pi_h(a' \mid s) = \frac{e^{\mathcal{T} h(s, a')}}{Z}$.
Using the quotient rule $\nabla_h (\frac{N}{D}) = \frac{(\nabla_h N)D - N(\nabla_h D)}{D^2}$:
Let $N_{a'} = e^{\mathcal{T} h(s,a')}$, so $\nabla_h N_{a'} = \mathcal{T} e^{\mathcal{T} h(s,a')} K((s,a'),\cdot)$.
Let $D = Z = \sum_{a''} e^{\mathcal{T} h(s,a'')}$, so $\nabla_h D = \mathcal{T} \sum_{a''} e^{\mathcal{T} h(s,a'')} K((s,a''),\cdot)$.
$$
\nabla_h \pi_h(a' \mid s) = \frac{\left(\mathcal{T} e^{\mathcal{T} h(s, a')} K((s, a'), \cdot)\right) Z - e^{\mathcal{T} h(s, a')} \left(\mathcal{T} \sum_{a''} e^{\mathcal{T} h(s, a'')} K((s, a''), \cdot)\right)}{Z^2}
$$
$$
= \mathcal{T} \frac{e^{\mathcal{T} h(s, a')}}{Z} K((s, a'), \cdot) - \mathcal{T} \frac{e^{\mathcal{T} h(s, a')}}{Z} \frac{\sum_{a''} e^{\mathcal{T} h(s, a'')} K((s, a''), \cdot)}{Z}
$$
$$
= \mathcal{T} \pi_h(a' \mid s) K((s, a'), \cdot) - \mathcal{T} \pi_h(a' \mid s) \sum_{a''} \pi_h(a'' \mid s) K((s, a''), \cdot)
$$
\begin{equation}
\nonumber
\nabla_h \pi_h(a' \mid s) = \mathcal{T} \pi_h(a' \mid s) \left( K((s, a'), \cdot) - \mathbb{E}_{a'' \sim \pi_h(\cdot \mid s)} \left[ K((s, a''), \cdot) \right] \right)
\end{equation}
Substituting this expression for $\nabla_h \pi_h(a' \mid s)$ back into the equation for $\nabla_h^2 \log \pi_h(a \mid s)$.
To align with the result in Lemma~\ref{theorem_RKHS_second} of the main paper, which states $\nabla_h^2 \log \pi_h(a_t \mid s_t) = -\mathcal{T} \operatorname{Cov}_{a' \sim \pi_h(\cdot \mid s_t)} [ K((s_t, a'), \cdot) ]$, the substitution effectively uses $\nabla_h \pi_h(a' \mid s) / \mathcal{T}$:
\begin{align} \label{eq:appendix_second_order_log_policy}
\nabla_h^2 \log \pi_h(a \mid s) &= -\mathcal{T} \sum_{a' \in \mathcal{A}} \pi_h(a' \mid s) \left( K((s, a'), \cdot) - \mathbb{E}_{a'' \sim \pi_h(\cdot \mid s)} \left[ K((s, a''), \cdot) \right] \right) \otimes K((s, a'), \cdot) \nonumber \\
&= -\mathcal{T} \left( \sum_{a' \in \mathcal{A}} \pi_h(a' \mid s) K((s, a'), \cdot) \otimes K((s, a'), \cdot) \right. \nonumber \\
& \quad \left. - \sum_{a' \in \mathcal{A}} \pi_h(a' \mid s) \left( \mathbb{E}_{a'' \sim \pi_h(\cdot \mid s)} \left[ K((s, a''), \cdot) \right] \right) \otimes K((s, a'), \cdot) \right) \nonumber \\
&= -\mathcal{T} \left( \mathbb{E}_{a' \sim \pi_h(\cdot \mid s)} \left[ K((s, a'), \cdot) \otimes K((s, a'), \cdot) \right] \right. \nonumber \\
& \quad \left. - \left( \mathbb{E}_{a'' \sim \pi_h(\cdot \mid s)} \left[ K((s, a''), \cdot) \right] \right) \otimes \left( \sum_{a' \in \mathcal{A}} \pi_h(a' \mid s) K((s, a'), \cdot) \right) \right) \nonumber \\
&= -\mathcal{T} \left( \mathbb{E}_{a' \sim \pi_h(\cdot \mid s)} \left[ K((s, a'), \cdot) \otimes K((s, a'), \cdot) \right] \right. \nonumber \\
& \quad \left. - \mathbb{E}_{a' \sim \pi_h(\cdot \mid s)} \left[ K((s, a'), \cdot) \right] \otimes \mathbb{E}_{a'' \sim \pi_h(\cdot \mid s)} \left[ K((s, a''), \cdot) \right] \right)
\end{align}
This can be compactly written using the covariance operator as defined in Lemma~\ref{theorem_RKHS_second} (using $s_t, a_t$ for generality):
\begin{equation}\nonumber
\nabla_h^2 \log \pi_h(a_t \mid s_t) = -\mathcal{T} \operatorname{Cov}_{a' \sim \pi_h(\cdot \mid s_t)} \left[ K((s_t, a'), \cdot) \right]
\end{equation}
This expression for $\nabla_h^2 \log \pi_h(a_t \mid s_t)$ is the component used in constructing the Hessian operator in Lemma~\ref{theorem_RKHS_second} and the estimated Hessian $\nabla_{h}^{2} \hat{J}\left(h_{k}\right)$ in the paper. By substituting the first and second derivative in \ref{Policy_Newton_formula} with $\nabla_h \log \pi_h(a_t \mid s_t)$ and $\nabla_h^2 \log \pi_h(a_t \mid s_t)$, the Lemma~\ref{theorem_RKHS_second} is proved.

\section{The derivation of Theorem \ref{theorem_final}}
\label{Derivation_theorem_final}
Theorem \ref{theorem_final} transforms the RKHS optimization problem for the Newton step $\Delta h$ (Equation~\eqref{RKHS_Auxiliary}) into an equivalent finite-dimensional optimization problem (Equation~\eqref{optimization_final}). The RKHS update step is $\Delta h = \bar{h}_{\boldsymbol{\alpha}}(\cdot) = \sum_{k=1}^{NT} \alpha_k K(x_k, \cdot)$, where $x_k = (s_k, a_k)$ are state-action pairs from the $N \times T$ trajectory data points ( $k$ is a flattened index from $1$ to $M=NT$), and $\boldsymbol{\bar{\alpha}} \in \mathbb{R}^{NT}$ is the coefficient vector.

The objective function in Equation~\eqref{RKHS_Auxiliary} is:
\begin{equation*}
L(\boldsymbol{\bar{\alpha}}) = \left\langle\nabla_{h} \hat{J}\left(h_{k}\right), \bar{h}_{\boldsymbol{\alpha}}\right\rangle + \frac{1}{2}\left\langle\nabla_{h}^{2} \hat{J} \left(h_{k}\right)\circ \bar{h}_{\boldsymbol{\alpha}}, \bar{h}_{\boldsymbol{\alpha}}\right\rangle+\frac{\beta}{6}\left\|\boldsymbol{\bar{\alpha}}\right\|_2^{3}
\end{equation*}
We derive the forms for the first two terms. The third term, $\frac{\beta}{6}\left\|\boldsymbol{\bar{\alpha}}\right\|_2^{3}$, directly uses the Euclidean norm of $\boldsymbol{\bar{\alpha}}$ as stated in Equation~\eqref{optimization_final}.

Let $M=NT$. The set of basis functions is $\{K(x_k, \cdot)\}_{k=1}^M$. The perturbation is $\bar{h}_{\boldsymbol{\alpha}}(\cdot) = \sum_{i=1}^{M} \alpha_i K(x_i, \cdot)$. We use index $i$ (or $j$) for the coefficients $\alpha_i$ and the basis functions $K(x_i, \cdot)$. We use index $l$ (or $l'$) for data points from the batch of $M$ samples when defining the gradient and Hessian operators.

\subsection{First-Order Term: $\left\langle\nabla_{h} \hat{J}\left(h_{k}\right), \bar{h}_{\boldsymbol{\alpha}}\right\rangle$}
The estimated first-order Fréchet derivative $\nabla_{h} \hat{J}(h_k)$ (denoted $g_{op}$ for operator form) is given by adapting Equation~\eqref{first_order_RKHS} for the empirical average over $N$ trajectories, or $M=NT$ total samples:
\begin{equation*}
g_{op}(\cdot) = \nabla_{h} \hat{J}(h_k)(\cdot) = \frac{\mathcal{T}}{N} \sum_{l=1}^{M} \Psi_l(\tau) \left( K(x_l, \cdot) - \mathbb{E}_{a' \sim \pi(\cdot \mid s_l)} \left[ K((s_l, a'), \cdot) \right] \right)
\end{equation*}
where $x_l=(s_l,a_l)$ is the $l$-th data point in the batch, and $\Psi_l(\tau)$ is its associated cumulative reward.
The inner product is:
\begin{align*}
\left\langle \nabla_{h} \hat{J}(h_k), \bar{h}_{\boldsymbol{\alpha}}\right\rangle &= \left\langle \nabla_{h} \hat{J}(h_k), \sum_{i=1}^{M} \alpha_i K(x_i, \cdot) \right\rangle \\
&= \sum_{i=1}^{M} \alpha_i \left\langle \nabla_{h} \hat{J}(h_k), K(x_i, \cdot) \right\rangle \quad \text{(by linearity of inner product)}
\end{align*}
Let $v_i = \left\langle \nabla_{h} \hat{J}(h_k), K(x_i, \cdot) \right\rangle$. Using the reproducing property and the expression for $\nabla_{h} \hat{J}(h_k)$:
\begin{align*}
v_i &= \frac{\mathcal{T}}{N} \sum_{l=1}^{M} \Psi_l(\tau) \left\langle K(x_l, \cdot) - \mathbb{E}_{a' \sim \pi(\cdot \mid s_l)} \left[ K((s_l, a'), \cdot) \right], K(x_i, \cdot) \right\rangle \\
&= \frac{\mathcal{T}}{N} \sum_{l=1}^{M} \Psi_l(\tau) \left( K(x_l, x_i) - \mathbb{E}_{a' \sim \pi(\cdot \mid s_l)} \left[ K((s_l, a'), x_i) \right] \right)
\end{align*}
Thus, $\left\langle \nabla_{h} \hat{J}(h_k), \bar{h}_{\boldsymbol{\alpha}}\right\rangle = \sum_{i=1}^{M} \alpha_i v_i = v^\top \boldsymbol{\bar{\alpha}}$. This definition of $v_i$ matches Theorem~\ref{theorem_final}.

\subsection{Second-Order Term: $\frac{1}{2}\left\langle\nabla_{h}^{2} \hat{J} \left(h_{k}\right)\circ \bar{h}_{\boldsymbol{\alpha}}, \bar{h}_{\boldsymbol{\alpha}}\right\rangle$}
Let $H_{op}^{est} = \nabla_{h}^{2} \hat{J}(h_k)$. Based on the updated Lemma~\ref{theorem_RKHS_second} and Equation~\eqref{Policy_Newton_formula}, the estimated Hessian operator $\nabla_h^2 \hat{J}(h_k)$ includes two components:
\begin{equation*}
 \frac{1}{N} \left[ \left( \sum_{l=1}^{M} \Psi_l(\tau) \nabla_h \log \pi_h(x_l)\right) \otimes \left( \sum_{l'=1}^{M} \nabla_h \log \pi_h(x_{l'}) \right) - \sum_{l=1}^{M} \Psi_l(\tau) \mathcal{T} \operatorname{Cov}_{a' \sim \pi(\cdot \mid s_l)} \left[ K(x_l, a') \right] \right]
\end{equation*}
Let $H_{op}^{(1)}$ be the operator for the first part (outer product) and $H_{op}^{(2)}$ for the second part (covariance sum), such that $\nabla_h^2 \hat{J}(h_k) = \frac{1}{N} (H_{op}^{(1)} - H_{op}^{(2)})$.

\textbf{1. Contribution from $H_{op}^{(1)}$:}
Let $\nabla_h \log \pi_h(x_l)(\cdot) = \mathcal{T} \left( K(x_l, \cdot) - \mathbb{E}_{a' \sim \pi(\cdot \mid s_l)} \left[ K((s_l, a'), \cdot) \right] \right)$.
Let $X_l(\bar{h}_{\boldsymbol{\alpha}}) = \langle \nabla_h \log \pi_h(x_l), \bar{h}_{\boldsymbol{\alpha}} \rangle$.
\begin{align*}
X_l(\bar{h}_{\boldsymbol{\alpha}}) &= \mathcal{T} \sum_{i=1}^{M} \alpha_i \left( K(x_l, x_i) - \mathbb{E}_{a' \sim \pi(\cdot \mid s_l)} \left[ K((s_l, a'), x_i) \right] \right)
\end{align*}
The quadratic form from $H_{op}^{(1)}$ is $\langle H_{op}^{(1)} \circ \bar{h}_{\boldsymbol{\alpha}}, \bar{h}_{\boldsymbol{\alpha}} \rangle = \left( \sum_{l=1}^{M} \Psi_l(\tau) X_l(\bar{h}_{\boldsymbol{\alpha}}) \right) \left( \sum_{l'=1}^{M} X_{l'}(\bar{h}_{\boldsymbol{\alpha}}) \right)$.
Using the definitions of $b_i$ and $c_i$ from Theorem~\ref{theorem_final} (with $K_{il} = K(x_i, x_l)$ and $K_{il}' = K(x_i, (s_l,a'))$, and summation index $l$ for data points):
\begin{align*}
b_i &= \sum_{l=1}^{M} \Psi_l(\tau) \left( K(x_i, x_l) - \mathbb{E}_{a' \sim \pi(\cdot \mid s_l)} \left[ K(x_i, (s_l, a')) \right] \right) \\
c_i &= \sum_{l=1}^{M} \left( K(x_i, x_l) - \mathbb{E}_{a' \sim \pi(\cdot \mid s_l)} \left[ K(x_i, (s_l, a')) \right] \right)
\end{align*}
Then,
$\sum_{l=1}^{M} \Psi_l(\tau) X_l(\bar{h}_{\boldsymbol{\alpha}}) = \mathcal{T} \sum_{i=1}^{M} \alpha_i b_i = \mathcal{T} (\boldsymbol{\bar{\alpha}}^\top b)$.
And,
$\sum_{l'=1}^{M} X_{l'}(\bar{h}_{\boldsymbol{\alpha}}) = \mathcal{T} \sum_{j=1}^{M} \alpha_j c_j = \mathcal{T} (\boldsymbol{\bar{\alpha}}^\top c)$.
So, the contribution to $\left\langle \nabla_h^2 \hat{J}(h_k) \circ \bar{h}_{\boldsymbol{\alpha}}, \bar{h}_{\boldsymbol{\alpha}} \right\rangle$ from this first part is:
\begin{equation*}
\frac{1}{N} \langle H_{op}^{(1)} \circ \bar{h}_{\boldsymbol{\alpha}}, \bar{h}_{\boldsymbol{\alpha}} \rangle = \frac{1}{N} (\mathcal{T} \boldsymbol{\bar{\alpha}}^\top b) (\mathcal{T} c^\top \boldsymbol{\bar{\alpha}}) = \boldsymbol{\bar{\alpha}}^\top \left( \frac{\mathcal{T}^2}{N} b c^\top \right) \boldsymbol{\bar{\alpha}}
\end{equation*}

\textbf{2. Contribution from $H_{op}^{(2)}$:}
$H_{op}^{(2)} \circ u = \sum_{l=1}^{M} \Psi_l(\tau) \mathcal{T} \operatorname{Cov}_{a' \sim \pi(\cdot \mid s_l)} [K((s_l,a'),\cdot)] \circ u$.
The quadratic form is $\langle H_{op}^{(2)} \circ \bar{h}_{\boldsymbol{\alpha}}, \bar{h}_{\boldsymbol{\alpha}} \rangle = \sum_{l=1}^{M} \Psi_l(\tau) \mathcal{T} \left\langle \operatorname{Cov}_{a' \sim \pi(\cdot \mid s_l)} [K((s_l,a'),\cdot)] \circ \bar{h}_{\boldsymbol{\alpha}}, \bar{h}_{\boldsymbol{\alpha}} \right\rangle$.
The inner term is $\operatorname{Var}_{a' \sim \pi(\cdot \mid s_l)} [\langle K((s_l,a'),\cdot), \bar{h}_{\boldsymbol{\alpha}} \rangle ] = \boldsymbol{\bar{\alpha}}^\top \Sigma^{(l)} \boldsymbol{\bar{\alpha}}$, where
$\Sigma^{(l)}_{ij} = \operatorname{Cov}_{a' \sim \pi(\cdot \mid s_l)} \left[ K((s_l,a'), x_i), K((s_l,a'), x_j) \right]$. Using $K_{il}' = K(x_i, (s_l,a'))$, this is $\Sigma^{(l)}_{ij} = \operatorname{Cov}_{a' \sim \pi(\cdot \mid s_l)} \left[ K_{il}', K_{jl}' \right]$, matching Theorem~\ref{theorem_final}.
So, the contribution to $\left\langle \nabla_h^2 \hat{J}(h_k) \circ \bar{h}_{\boldsymbol{\alpha}}, \bar{h}_{\boldsymbol{\alpha}} \right\rangle$ from this second part is:
\begin{equation*}
-\frac{1}{N} \langle H_{op}^{(2)} \circ \bar{h}_{\boldsymbol{\alpha}}, \bar{h}_{\boldsymbol{\alpha}} \rangle = -\frac{1}{N} \boldsymbol{\bar{\alpha}}^\top \left( \mathcal{T} \sum_{l=1}^{M} \Psi_l(\tau) \Sigma^{(l)} \right) \boldsymbol{\bar{\alpha}} = \boldsymbol{\bar{\alpha}}^\top \left( - \frac{\mathcal{T}}{N} \sum_{l=1}^{M} \Psi_l(\tau) \Sigma^{(l)} \right) \boldsymbol{\bar{\alpha}}
\end{equation*}

\textbf{Combining terms for the matrix $H$:}
The full quadratic form for the second-order term in the objective function is $\frac{1}{2} \boldsymbol{\bar{\alpha}}^\top H \boldsymbol{\bar{\alpha}}$, where the matrix $H$ is given by:
\begin{equation*}
H = \frac{\mathcal{T}^2}{N} b c^\top - \frac{\mathcal{T}}{N} \sum_{l=1}^{M} \Psi_l(\tau) \Sigma^{(l)}
\end{equation*}
This matches Equation~\eqref{eq:H_definition} in Theorem~\ref{theorem_final}.

\section{The proof of the convergence}
\subsection{Proof for Monte Carlo convergence}
\label{MC_convergence}
Let the Monte Carlo estimates be defined as the average of $N$ independent and identically distributed (i.i.d.) samples, denoted by $g_h^{(i)}$ and $H_h^{(i)}$, corresponding to trajectories $\tau_i \sim p(\tau; \pi)$.
\begin{equation}
\nonumber
    \nabla_{h} \hat{J}\left(h_{k}\right) = \frac{1}{N} \sum_{i=1}^{N} g_h^{(i)} 
\end{equation}
\begin{equation}
\nonumber
    \nabla_{h}^2 \hat{J}\left(h_{k}\right) = \frac{1}{N} \sum_{i=1}^{N} H_h^{(i)}
\end{equation}
The true Fréchet derivatives are the expectations of these samples (we use $\mathbb{E}[\cdot]$ as shorthand for $\mathbb{E}_{\tau \sim p(\tau ; \pi)}[\cdot]$):
\begin{equation}
\nonumber
    \nabla_{h} J\left(h_{k}\right) = \mathbb{E}[g_h] 
\end{equation}
\begin{equation}
\nonumber
    \nabla_{h}^2 J\left(h_{k}\right) = \mathbb{E}[H_h]
\end{equation}

Consider the expected squared norm for the first-order derivative estimate. Since the Monte Carlo estimator is unbiased ($\mathbb{E}[\nabla_{h} \hat{J}(h_{k})] = \nabla_{h} J(h_{k})$), the expected squared norm equals the variance:
\begin{equation}
\nonumber
    \mathbb{E} \left[\norm{\nabla_{h} \hat{J}\left(h_{k}\right)-\nabla_{h} J\left(h_{k}\right)}^2 \right] = \text{Var}(\nabla_{h} \hat{J}(h_{k})) 
\end{equation}
The variance of the mean of $N$ random variables is $1/N$ times the variance of a single variable:
\begin{equation}
\nonumber
    \text{Var}(\nabla_{h} \hat{J}(h_{k})) = \text{Var}\left(\frac{1}{N} \sum_{i=1}^{N} g_h^{(i)}\right) = \frac{1}{N^2} \sum_{i=1}^{N} \text{Var}(g_h^{(i)}) = \frac{1}{N} \text{Var}(g_h) 
\end{equation}
The variance of $g_h$ is defined as:
\begin{equation}
\nonumber
    \text{Var}(g_h) = \mathbb{E}[\norm{g_h - \mathbb{E}[g_h]}^2] 
\end{equation}
Using the property $\text{Var}(X) = \mathbb{E}[||X||^2] - ||\mathbb{E}[X]||^2$ (which holds according to the definition of RKHS) and the fact that $||\mathbb{E}[X]||^2 \ge 0$:
\begin{equation}
\nonumber
    \text{Var}(g_h) = \mathbb{E}[\norm{g_h}^2] - \norm{\mathbb{E}[g_h]}^2 \leq \mathbb{E}[\norm{g_h}^2] 
\end{equation}
Applying the assumption $\mathbb{E}[\norm{g_h}^2] \leq \sigma^2_{0}$:
\begin{equation}
\nonumber
    \text{Var}(g_h) \leq \sigma^2_{0} 
\end{equation}
Substituting this back, we find the standard convergence rate for the expected squared norm:
\begin{equation}
\nonumber
    \mathbb{E} \left[\norm{\nabla_{h} \hat{J}\left(h_{k}\right)-\nabla_{h} J\left(h_{k}\right)}^2 \right] \leq \frac{\sigma^2_{0}}{N} 
\end{equation}

Similarly, for the second-order derivative estimate:
\begin{equation}
\nonumber
    \mathbb{E} \left[\norm{\nabla_{h}^2 \hat{J}\left(h_{k}\right)-\nabla_{h}^2 J\left(h_{k}\right)}^2 \right] = \text{Var}(\nabla_{h}^2 \hat{J}(h_{k})) 
\end{equation}
\begin{equation}
\nonumber
    \text{Var}(\nabla_{h}^2 \hat{J}(h_{k})) = \text{Var}\left(\frac{1}{N} \sum_{i=1}^{N} H_h^{(i)}\right) = \frac{1}{N} \text{Var}(H_h) 
\end{equation}
\begin{equation}
\nonumber
    \text{Var}(H_h) = \mathbb{E}[\norm{H_h - \mathbb{E}[H_h]}^2] \leq \mathbb{E}[\norm{H_h}^2] 
\end{equation}
Applying the assumption $\mathbb{E}[\norm{H_h}^2] \leq \sigma^2_{1}$:
\begin{equation}
\nonumber
    \text{Var}(H_h) \leq \sigma^2_{1} 
\end{equation}
Substituting back, the standard convergence rate for the expected squared norm is:
\begin{equation}
\nonumber
    \mathbb{E} \left[\norm{\nabla_{h}^2 \hat{J}\left(h_{k}\right)-\nabla_{h}^2 J\left(h_{k}\right)}^2 \right] \leq \frac{\sigma^2_{1}}{N} 
\end{equation}

This completes the proof.

\subsection{Proof for Taylor upper bound}
\label{Taylor upper bound}
Let $\Delta h = h_2 - h_1$. Define the auxiliary function $\phi: [0, 1] \to \mathbb{R}$ by $\phi(t) = J(h_1 + t \Delta h)$.
By the chain rule for Fr\'echet derivatives:
\begin{equation}\nonumber
\phi'(t) = \langle \nabla_h J(h_1 + t \Delta h), \Delta h \rangle
\end{equation}
\begin{equation}\nonumber
\phi''(t) = \langle \nabla_h^2 J(h_1 + t \Delta h) \circ \Delta h, \Delta h \rangle
\end{equation}
Using Taylor's theorem with integral remainder for $\phi(t)$:
\begin{equation}\nonumber
\phi(1) = \phi(0) + \phi'(0) + \int_0^1 (1-t) \phi''(t) dt
\end{equation}
Substituting the expressions for $\phi$, $\phi'$, and $\phi''$:
\begin{equation}\nonumber
J(h_2) = J(h_1) + \langle \nabla_h J(h_1), \Delta h \rangle + \int_0^1 (1-t) \langle \nabla_h^2 J(h_1 + t \Delta h) \circ \Delta h, \Delta h \rangle dt
\end{equation}
We introduce the second-order term at $h_1$. Note that $\int_0^1 (1-t) dt = 1/2$. Thus,
\begin{equation}\nonumber
\frac{1}{2} \langle \nabla_h^2 J(h_1) \circ \Delta h, \Delta h \rangle = \int_0^1 (1-t) \langle \nabla_h^2 J(h_1) \circ \Delta h, \Delta h \rangle dt
\end{equation}
Adding and subtracting this term within the integral expression for $J(h_2)$:
\begin{align*}\nonumber
J(h_2) &= J(h_1) + \langle \nabla_h J(h_1), \Delta h \rangle + \frac{1}{2} \langle \nabla_h^2 J(h_1) \circ \Delta h, \Delta h \rangle \\
& \quad + \int_0^1 (1-t) \langle \nabla_h^2 J(h_1 + t \Delta h) \circ \Delta h, \Delta h \rangle dt \\
& \quad - \int_0^1 (1-t) \langle \nabla_h^2 J(h_1) \circ \Delta h, \Delta h \rangle dt \\
&= J(h_1) + \langle \nabla_h J(h_1), \Delta h \rangle + \frac{1}{2} \langle \nabla_h^2 J(h_1)\circ \Delta h, \Delta h \rangle \\
& \quad + \int_0^1 (1-t) \langle [\nabla_h^2 J(h_1 + t \Delta h) - \nabla_h^2 J(h_1)] \circ \Delta h, \Delta h \rangle dt
\end{align*}
Let $R_2$ be the remainder term:
\begin{equation}\nonumber
R_2 = \int_0^1 (1-t) \langle [\nabla_h^2 J(h_1 + t \Delta h) - \nabla_h^2 J(h_1)] \circ \Delta h, \Delta h \rangle dt
\end{equation}
Through Cauchy-Schwarz inequality, the quadratic form $\langle A v, v \rangle \le \|A\| \|v\|^2$:
\begin{equation}\nonumber
\langle [\nabla_h^2 J(h_1 + t \Delta h) - \nabla_h^2 J(h_1)] \circ \Delta h, \Delta h \rangle \le \| \nabla_h^2 J(h_1 + t \Delta h) - \nabla_h^2 J(h_1) \| \|\Delta h\|^2
\end{equation}
Using the Lipschitz continuity of the Hessian in Assumption \ref{Lipschitz}:
\begin{equation}\nonumber
\| \nabla_h^2 J(h_1 + t \Delta h) - \nabla_h^2 J(h_1) \| \le L \| (h_1 + t \Delta h) - h_1 \| = L \| t \Delta h \| = L t \|\Delta h\| \quad (\text{since } t \ge 0)
\end{equation}
Substituting this bound into the integral for $R_2$:
\begin{align*}\nonumber
R_2 &\le \int_0^1 (1-t) (L t \|\Delta h\|) \|\Delta h\|^2 dt \\
&= L \|\Delta h\|^3 \int_0^1 (1-t) t dt \\
&= L \|\Delta h\|^3 \int_0^1 (t - t^2) dt \\
&= \frac{L}{6} \|\Delta h\|^3
\end{align*}
Substituting this upper bound for $R_2$ back into the expression for $J(h_2)$ and replacing $\Delta h$ with $h_2 - h_1$:
\begin{equation}\nonumber
J(h_2) \le J(h_1) + \langle \nabla_h J(h_1), h_2 - h_1 \rangle + \frac{1}{2} \langle \nabla_h^2 J(h_1) \circ(h_2 - h_1), h_2 - h_1 \rangle + \frac{L}{6} \|h_2 - h_1\|^3
\end{equation}
This completes the proof.

\subsection{Proof for the upper bound of the step norm}
\label{proof_step}
To prove this upper bound, we first need to introduce a lemma to show the optimality conditions of the iteration step:

\begin{lemma}
\label{Optimality}
    (Optimality conditions) Let $$\Delta h = \underset{\bar{h} \in \mathcal{H}_{K}}{\operatorname{argmin}}\left\{\left\langle\nabla_{h} \hat{J}\left(h_{k}\right), \bar{h}\right\rangle \\
+\frac{1}{2}\left\langle\nabla_{h}^{2} \hat{J} \left(h_{k}\right)\circ \bar{h}, \bar{h}\right\rangle+\frac{\beta}{6}\left\|\bar{h}\right\|^{3}\right\}$$, then it satisfies that:
\begin{equation}
\nonumber
\begin{aligned}
    &\nabla_{h} \hat{J}\left(h_{k}\right) + \nabla_{h}^{2} \hat{J} \left(h_{k}\right) \circ \Delta h + \frac{\beta}{2}\|\Delta h\| \Delta h = 0 \text{ (necessary condition),}\\
    &\innerprod{(\nabla_{h}^{2} \hat{J} \circ u}{u} + \innerprod{ \frac{\beta}{2}\|\Delta h\| I \circ u}{u} \ge 0 \; \forall u \in \mathcal{H}_{K}  \text{ (sufficient condition),}
\end{aligned}
\end{equation}
\end{lemma}
where $I$ is the identity operator on $\mathcal{H}_{K} \otimes \mathcal{H}_{K} \to \mathcal{H}_{K}$.

\textbf{Proof:}
To simplify the notation in the proof, we denote $g = \nabla J(h_k) \in \mathcal{H}_{K}$, $H_{k} = \nabla^2 J(h_k)$ and the objective function $M: \mathcal{H}_{K} \to \mathbb{R}$:
\begin{equation}\nonumber
M(\bar{h}) = \innerprod{g}{\bar{h}} + \frac{1}{2}\innerprod{(H_{k} \circ \bar{h})}{\bar{h}} + \frac{\beta}{6}\norm{\bar{h}}^{3}.
\end{equation}

The first Fr\'echet derivative of $M$ at $\bar{h}$ is given by:
\begin{equation}\nonumber
\nabla M(\bar{h}) = g + H_{k} \circ \bar{h} + \frac{\beta}{2}\norm{\bar{h}}\bar{h}
\end{equation}
Since $\Delta h$ is a minimizer, it must satisfy the necessary condition $\nabla M(\Delta h) = 0$:
\begin{equation}\nonumber
g + H_{k} \circ \Delta h + \frac{\beta}{2}\norm{\Delta h}\Delta h = 0.
\end{equation}

We now prove the standard second-order necessary condition for $\Delta h$.

Let $\Delta h$ be a minimizer of $M(\bar{h})$ and let $\frac{\beta}{2}\|\Delta h\| = \frac{\beta}{2}\norm{\Delta h}$. Then the operator $H_{k} + \frac{\beta}{2}\|\Delta h\| I$ must be positive semi-definite, i.e.,
\begin{equation}\nonumber
\innerprod{(H_{k} \circ u}{u} + \innerprod{ \frac{\beta}{2}\|\Delta h\| I \circ u}{u} \ge 0 \;  \to H_{k} + \frac{\beta}{2}\|\Delta h\| I \succeq 0
\end{equation}

We proceed by contradiction. Assume that $H_{k} + \frac{\beta}{2}\|\Delta h\| I$ is not positive semi-definite ($H_{k} + \frac{\beta}{2}\|\Delta h\| I \not\succeq 0$). This implies that there exists a direction $u \in \mathcal{H}_{K}$ with $\norm{u}=1$ such that its associated quadratic form is negative:
\begin{equation}\nonumber
\mu := \innerprod{(H_{k} + \frac{\beta}{2}\|\Delta h\| I) \circ u}{u} < 0
\end{equation}

Consider the Taylor expansion of $M$ around the minimizer $\Delta h$ along the direction $u$ for a small step $\epsilon \in \mathbb{R}$. Using Taylor's theorem in Hilbert spaces:
\begin{equation}\nonumber
M(\Delta h + \epsilon u) = M(\Delta h) + \epsilon \innerprod{\nabla M(\Delta h)}{u} + \frac{\epsilon^2}{2} \innerprod{(\nabla^2 M(\Delta h)) \circ u}{u} + O(\epsilon^3)
\end{equation}
Since $\nabla M(\Delta h) = 0$ from the first-order condition, this simplifies to:
\begin{equation}\nonumber
M(\Delta h + \epsilon u) = M(\Delta h) + \frac{\epsilon^2}{2} \innerprod{(\nabla^2 M(\Delta h)) \circ u}{u} + O(\epsilon^3)
\end{equation}
The second Fr\'echet derivative (Hessian operator) of $M$ at $\bar{h}$ is calculated as:
\begin{equation}\nonumber
\nabla^2 M(\bar{h}) = H_{k} + \frac{\beta}{2} \frac{\bar{h} \otimes \bar{h}}{\norm{\bar{h}}} + \frac{\beta}{2} \norm{\bar{h}} I
\end{equation}
Evaluating at $\Delta h$ (assuming $\Delta h \neq 0$, which implies $\frac{\beta}{2}\|\Delta h\| > 0$; the case $\Delta h = 0$ requires separate, simpler verification) and substituting $\frac{\beta}{2}\|\Delta h\| = \frac{\beta}{2}\norm{\Delta h}$ yields:
\begin{equation}\nonumber
\nabla^2 M(\Delta h) = H_{k} + \frac{\beta}{2} \frac{\Delta h \otimes \Delta h}{\norm{\Delta h}} + \frac{\beta}{2}\|\Delta h\| I = (H_{k} + \frac{\beta}{2}\|\Delta h\| I) + \frac{\beta}{2} \frac{\Delta h \otimes \Delta h}{\norm{\Delta h}}
\end{equation}
Now, substitute this Hessian back into the Taylor expansion. The quadratic term is:
\begin{equation}\nonumber
\innerprod{(\nabla^2 M(\Delta h)) \circ u}{u} = \innerprod{(H_{k} + \frac{\beta}{2}\|\Delta h\| I) \circ u}{u} + \frac{\beta}{2 \norm{\Delta h}} \innerprod{\innerprod{\Delta h}{u} \Delta h}{u}
\end{equation}
Using the definition of $\mu$ and properties of the inner product, this becomes:
\begin{equation}\nonumber
\innerprod{(\nabla^2 M(\Delta h)) \circ u}{u} = \mu + \frac{\beta}{2 \norm{\Delta h}} \innerprod{\Delta h}{u}^2
\end{equation}
The Taylor expansion for the difference is thus:
\begin{equation}\nonumber
M(\Delta h + \epsilon u) - M(\Delta h) = \frac{\epsilon^2}{2} \left( \mu + \frac{\beta}{2 \norm{\Delta h}} \innerprod{\Delta h}{u}^2 \right) + O(\epsilon^3)
\end{equation}
Let $K = \mu + \frac{\beta}{2 \norm{\Delta h}} \innerprod{\Delta h}{u}^2$. By assumption, $\mu < 0$. The second term $\frac{\beta}{2 \norm{\Delta h}} \innerprod{\Delta h}{u}^2$ is non-negative. Since the assumption $H_{k} + \frac{\beta}{2}\|\Delta h\| I \not\succeq 0$ leads to a contradiction in all cases, the assumption must be false. Therefore, we must conclude that $H_{k} + \frac{\beta}{2}\|\Delta h\| I \succeq 0$, which complete the proof.

Now we continue the proof for the upper bound of the step norm. Through the Taylor upper bound in lemma \ref{Taylor_uppper} we know that:
\begin{equation}\nonumber
J(h_2) \le J(h_1) + \langle \nabla_h J(h_1), h_2 - h_1 \rangle + \frac{1}{2} \langle \nabla_h^2 J(h_1) \circ (h_2 - h_1), h_2 - h_1 \rangle + \frac{L}{6} \|h_2 - h_1\|^3.
\end{equation}
For $\Delta h = h_2 - h_1$ that satisfies the optimality conditions, we can use the necessary condition to establish that:
\begin{equation}
\label{Intermid_inequ}
\begin{aligned}
    J(h_2) \leq &J(h_1) + \langle \nabla_h J(h_1) - \nabla_{h} \hat{J}\left(h_{k}\right), \Delta h \rangle + \frac{1}{2} \langle \left(\nabla_h^2 J(h_1) - \nabla_h^2 \hat{J}(h_1) \right)\circ (\Delta h), \Delta h \rangle + \\
    &\frac{L}{6} \|\Delta h\|^3 - \frac{1}{2} \langle \left(\nabla_h^2 \hat{J}(h_1) \right)\circ (\Delta h), \Delta h \rangle - \frac{\beta}{2}\|\Delta h\|^3.
\end{aligned}
\end{equation}
For $L \leq \beta$ and the sufficient condition in Lemma \ref{Optimality}, we could find that

\begin{gather}
\nonumber
    \frac{L}{6} \|\Delta h\|^3 \leq \frac{\beta}{6} \|\Delta h\|^3 \\
    \nonumber
    - \frac{1}{2} \langle \left(\nabla_h^2 \hat{J}(h_1) \right)\circ (\Delta h), \Delta h \rangle \leq \frac{\beta}{4} \|\Delta h\|^3
\end{gather}
Substituting back into the inequality~\eqref{Intermid_inequ}, we can find that:
\begin{equation}
\label{Initial_inequ}
\begin{aligned}
    &\frac{\beta}{12}\|\Delta h\|^3 \leq J(h_1) -J(h_2) \\  &+ \langle \nabla_h J(h_1) - \nabla_{h} \hat{J}\left(h_{k}\right), \Delta h \rangle + \frac{1}{2} \langle \left(\nabla_h^2 J(h_1) - \nabla_h^2 \hat{J}(h_1) \right)\circ (\Delta h), \Delta h \rangle 
\end{aligned}
\end{equation}

For the gradient error term, applying Cauchy-Schwarz and Young's inequality ($ab \leq C_1 a^{3/2} + \epsilon b^3$ with $a=\|\nabla_h J(h_k) - \nabla_{h} \hat{J}(h_k)\|$, $b=\|\Delta h_k\|$, $p=3/2$, $q=3$, and $\epsilon = \beta/36$):
\begin{align*}
    &\langle \nabla_h J(h_k) - \nabla_{h} \hat{J}(h_k), \Delta h_k \rangle \\
    &\leq \|\nabla_h J(h_k) - \nabla_{h} \hat{J}(h_k)\| \|\Delta h_k\| \\
    &\leq \frac{4\sqrt{3}}{3\sqrt{\beta}} \|\nabla_h J(h_k) - \nabla_{h} \hat{J}(h_k)\|^{3/2} + \frac{\beta}{36} \|\Delta h_k\|^3.
\end{align*}
For the Hessian error term, applying generalized Cauchy-Schwarz and Young's inequality ($cd^2 \leq C_2 c^3 + \epsilon d^3$ with $c = \frac{1}{2}\|\nabla_h^2 J(h_k) - \nabla_h^2 \hat{J}(h_k)\|$, $d=\|\Delta h_k\|$, $p=3$, $q=3/2$, and $\epsilon = \beta/36$):
\begin{align*}
    &\frac{1}{2} \langle (\nabla_h^2 J(h_k) - \nabla_h^2 \hat{J}(h_k) )\circ (\Delta h_k), \Delta h_k \rangle \\
    &\leq \frac{1}{2} \|\nabla_h^2 J(h_k) - \nabla_h^2 \hat{J}(h_k)\| \|\Delta h_k\|^2 \\
    &\leq \frac{24}{\beta^2} \|\nabla_h^2 J(h_k) - \nabla_h^2 \hat{J}(h_k)\|^3 + \frac{\beta}{36} \|\Delta h_k\|^3.
\end{align*}
Substituting these bounds back into \eqref{Initial_inequ}:
\begin{align*}
    \frac{\beta}{12}\|\Delta h_k\|^3 \leq &J(h_k) -J(h_{k+1}) \\
    & + \frac{4\sqrt{3}}{3\sqrt{\beta}} \|\nabla_h J(h_k) - \nabla_{h} \hat{J}(h_k)\|^{3/2} + \frac{\beta}{36} \|\Delta h_k\|^3 \\
    & + \frac{24}{\beta^2} \|\nabla_h^2 J(h_k) - \nabla_h^2 \hat{J}(h_k)\|^3 + \frac{\beta}{36} \|\Delta h_k\|^3.
\end{align*}
Rearranging terms yields:
\begin{equation*}
\begin{aligned}
   &\left(\frac{\beta}{12} - \frac{\beta}{36} - \frac{\beta}{36}\right) \|\Delta h_k\|^3 \leq J(h_k) -J(h_{k+1}) + \\
   &\frac{4\sqrt{3}}{3\sqrt{\beta}} \|\nabla_h J(h_k) - \nabla_{h} \hat{J}(h_k)\|^{3/2} + \frac{24}{\beta^2} \|\nabla_h^2 J(h_k) - \nabla_h^2 \hat{J}(h_k)\|^3, 
\end{aligned}
\end{equation*}
which simplifies to:
\begin{equation} \label{eq:intermediate_bound_k_full}
\frac{\beta}{36} \|\Delta h_k\|^3 \leq J(h_k) -J(h_{k+1}) + \frac{4\sqrt{3}}{3\sqrt{\beta}} \|\nabla_h J(h_k) - \nabla_{h} \hat{J}(h_k)\|^{3/2} + \frac{24}{\beta^2} \|\nabla_h^2 J(h_k) - \nabla_h^2 \hat{J}(h_k)\|^3.
\end{equation}
Now, we take the total expectation $\mathbb{E}[\cdot]$ over all randomness. Using Lemma \ref{MC_convergence} and properties of expectation (Jensen's inequality), we bound the expected error terms:
\begin{align*}
\mathbb{E}\left[ \|\nabla_h J(h_k) - \nabla_{h} \hat{J}(h_k)\|^{3/2} \right] &\leq \left( \mathbb{E}\left[ \|\nabla_h J(h_k) - \nabla_{h} \hat{J}(h_k)\|^2 \right] \right)^{3/4} \leq \left( \frac{\sigma_0^2}{N} \right)^{3/4} = \frac{\sigma_0^{3/2}}{N^{3/4}}. \\
\mathbb{E}\left[ \|\nabla_h^2 J(h_k) - \nabla_h^2 \hat{J}(h_k)\|^3 \right] &\leq \left( \frac{\sigma_1}{\sqrt{N}} \right)^3 = \frac{\sigma_1^3}{N^{3/2}}. \quad \text{(See note below)}
\end{align*}

Substituting these bounds into the expectation of \eqref{eq:intermediate_bound_k_full}:
\begin{align*}
\frac{\beta}{36} \mathbb{E}\left[\|\Delta h_k\|^3\right] \leq &\mathbb{E}[J(h_k)] - \mathbb{E}[J(h_{k+1})] \\
&+ \frac{4\sqrt{3}}{3\sqrt{\beta}} \left( \frac{\sigma_0^{3/2}}{N^{3/4}} \right) + \frac{24}{\beta^2} \left( \frac{\sigma_1^3}{N^{3/2}} \right).
\end{align*}
Summing this inequality over the total iterations $k=1, \ldots, M$:
\begin{align*}
\frac{\beta}{36} \sum_{k=1}^{M} \mathbb{E}\left[\|h_{k+1}-h_k\|^3\right] \leq &\sum_{k=1}^{M} \left( \mathbb{E}[J(h_k)] - \mathbb{E}[J(h_{k+1})] \right) \\
& + \sum_{k=1}^{M} \left( \frac{4\sqrt{3}}{3\sqrt{\beta}} \frac{\sigma_0^{3/2}}{N^{3/4}} + \frac{24}{\beta^2} \frac{\sigma_1^3}{N^{3/2}} \right).
\end{align*}
The first sum on the right-hand side telescopes to $\mathbb{E}[J(h_1)] - \mathbb{E}[J(h_{M+1})]$. Assuming $h_1$ is deterministic and $J(h) \ge J^*$ for some minimum value $J^*$, this sum is bounded by $J(h_1) - J^*$. The second sum consists of terms independent of the summation index $k$:
$$
\sum_{k=1}^{M} \left( \dots \right) = M \left( \frac{4\sqrt{3}}{3\sqrt{\beta}} \frac{\sigma_0^{3/2}}{N^{3/4}} + \frac{24}{\beta^2} \frac{\sigma_1^3}{N^{3/2}} \right).
$$
Combining these results:
$$
\frac{\beta}{36} \sum_{k=1}^{M} \mathbb{E}\left[\|h_{k+1}-h_k\|^3\right] \leq J(h_1) - J^* + M \left( \frac{4\sqrt{3}}{3\sqrt{\beta}} \frac{\sigma_0^{3/2}}{N^{3/4}} + \frac{24}{\beta^2} \frac{\sigma_1^3}{N^{3/2}} \right).
$$
Let $R$ be a random variable uniformly distributed on $\{1, \ldots, M\}$, such that $P(R=k) = 1/M$. Then $\mathbb{E}[\|h_{R+1}-h_R\|^3] = \frac{1}{M} \sum_{k=1}^{M} \mathbb{E}\left[\|h_{k+1}-h_k\|^3\right]$. Dividing the inequality by $M$:
$$
\frac{\beta}{36} \mathbb{E}\left[\|h_{R+1}-h_R\|^3\right] \leq \frac{J(h_1) - J^*}{M} + \left( \frac{4\sqrt{3}}{3\sqrt{\beta}} \frac{\sigma_0^{3/2}}{N^{3/4}} + \frac{24}{\beta^2} \frac{\sigma_1^3}{N^{3/2}} \right).
$$
Finally, multiplying by $36/\beta$ isolates the expected cubic step norm for a randomly chosen iteration $R$:
\begin{align} \label{eq:final_step_norm_bound_corrected_M}
\mathbb{E}\left[\|h_{R+1}-h_R\|^3\right] &\leq \frac{36(J(h_1) - J^*)}{\beta M} + \frac{36}{\beta} \left( \frac{4\sqrt{3}}{3\sqrt{\beta}} \frac{\sigma_0^{3/2}}{N^{3/4}} + \frac{24}{\beta^2} \frac{\sigma_1^3}{N^{3/2}} \right) \nonumber \\
&\leq \frac{36(J(h_1) - J^*)}{\beta M} + \frac{48\sqrt{3}}{\beta^{3/2}} \frac{\sigma_0^{3/2}}{N^{3/4}} + \frac{864}{\beta^3} \frac{\sigma_1^3}{N^{3/2}}.
\end{align}
We can box the final result for emphasis:
\begin{center}
\begin{equation*} 
\mathbb{E}\left[\|h_{R+1}-h_R\|^3\right] \leq \frac{36(J(h_1) - J^*)}{\beta M} + \frac{48\sqrt{3}}{\beta^{3/2}} \frac{\sigma_0^{3/2}}{N^{3/4}} + \frac{864}{\beta^3} \frac{\sigma_1^3}{N^{3/2}}.
\end{equation*}
\end{center}
This completes the derivation of the upper bound on the expected cubic step norm.

\subsection{Proof for the lower bound of the step norm}
\label{proof_step_lower}
From the proof of the Taylor upper bound in Appendix \ref{Taylor upper bound}, we could similarly derive the first-order Taylor upper bound as
\begin{equation}
\label{first-order taylor}
    \nabla J(h_2) \le \nabla J(h_1) + \langle \nabla_h^2 J(h_1) \circ (h_2 - h_1), h_2 - h_1 \rangle + \frac{L}{2} \|h_2 - h_1\|^3.
\end{equation}
Through this, we could prove the lower bound by first constructing this auxiliary equation through the optimality conditions in Lemma \ref{Optimality}:
\begin{align*}
    \nabla J(h_{k+1}) &= \nabla J(h_{k+1}) - (\nabla_h \hat{J}(h_k) + \nabla_h^2 \hat{J}(h_k) \circ \Delta h_k + \frac{\beta}{2} \|\Delta h_k\| \Delta h_k) \\
    &= \left[ \nabla J(h_{k+1}) - \nabla J(h_k) - \nabla^2 J(h_k) \circ \Delta h_k \right] \quad \text{(Term 1)}\\
    & \quad + \left[ \nabla J(h_k) - \nabla_{h} \hat{J}(h_k) \right] \quad \text{(Term 2)}\\
    & \quad + \left[ (\nabla^2 J(h_k) - \nabla_{h}^2 \hat{J}(h_k)) \circ \Delta h_k \right] \quad \text{(Term 3)}\\
    & \quad - \frac{\beta}{2} \|\Delta h_k\| \Delta h_k \quad \text{(Term 4)}
\end{align*}
Taking norms and applying the triangle inequality:
\begin{align*}
    \|\nabla J(h_{k+1})\| \leq & \|\nabla J(h_{k+1}) - \nabla J(h_k) - \nabla^2 J(h_k) \circ \Delta h_k\| \\
    & + \|\nabla J(h_k) - \nabla_{h} \hat{J}(h_k)\| \\
    & + \|(\nabla^2 J(h_k) - \nabla_{h}^2 \hat{J}(h_k)) \circ \Delta h_k\| \\
    & + \| - \frac{\beta}{2} \|\| \Delta h_k \|\Delta h_k
\end{align*}
We bound the terms using Assumption \ref{Lipschitz} for Term 1, norm properties for Term 3, and direct calculation for Term 4:
\begin{align*}
 \|\nabla J(h_{k+1})\| \leq & \frac{L}{2} \|\Delta h_k\|^2 + \|\nabla J(h_k) - \nabla_{h} \hat{J}(h_k)\| \\
 & + \|\nabla^2 J(h_k) - \nabla_{h}^2 \hat{J}(h_k)\| \|\Delta h_k\| + \frac{\beta}{2} \|\Delta h_k\|^2
\end{align*}
Applying Young's inequality $ab \leq \frac{a^2}{2C} + \frac{C b^2}{2}$ with $C=L+\beta$ to the term involving the Hessian error:
\begin{align*}
 \|\nabla^2 J(h_k) - \nabla_{h}^2 \hat{J}(h_k)\| \|\Delta h_k\| \leq \frac{\|\nabla^2 J(h_k) - \nabla_{h}^2 \hat{J}(h_k)\|^2}{2(L+\beta)} + \frac{(L+\beta)\|\Delta h_k\|^2}{2}
\end{align*}
Substituting this back and collecting terms with $\|\Delta h_k\|^2$:
\begin{align*}
 \|\nabla J(h_{k+1})\| &\leq \left(\frac{L}{2} + \frac{L+\beta}{2} + \frac{\beta}{2} \right) \|\Delta h_k\|^2 \\
 & \quad + \|\nabla J(h_k) - \nabla_{h} \hat{J}(h_k)\| + \frac{\|\nabla^2 J(h_k) - \nabla_{h}^2 \hat{J}(h_k)\|^2}{2(L+\beta)} \\
 &= \left(L + \beta \right) \|\Delta h_k\|^2 \\
 & \quad + \|\nabla J(h_k) - \nabla_{h} \hat{J}(h_k)\| + \frac{\|\nabla^2 J(h_k) - \nabla_{h}^2 \hat{J}(h_k)\|^2}{2(L+\beta)}
\end{align*}
Now, take the total expectation $\mathbb{E}[\cdot]$. Using Lemma \ref{MC_convergence} and Jensen's inequality:
\begin{align*}
 \mathbb{E}\left[ \|\nabla J(h_k) - \nabla_{h} \hat{J}(h_k)\| \right] &\leq \sqrt{\mathbb{E}\left[ \|\nabla J(h_k) - \nabla_{h} \hat{J}(h_k)\|^2 \right]} \leq \frac{\sigma_0}{\sqrt{N}} \\
 \mathbb{E}\left[ \|\nabla^2 J(h_k) - \nabla_{h}^2 \hat{J}(h_k)\|^2 \right] &\leq \frac{\sigma_1^2}{N}
\end{align*}
Applying expectation to the inequality for $\|\nabla J(h_{k+1})\|$:
\begin{align*}
 \mathbb{E}[\|\nabla J(h_{k+1})\|] &\leq \left(L + \beta \right) \mathbb{E}[\|\Delta h_k\|^2] \\
 &\quad + \mathbb{E}\left[\|\nabla J(h_k) - \nabla_{h} \hat{J}(h_k)\|\right] + \frac{\mathbb{E}\left[\|\nabla^2 J(h_k) - \nabla_{h}^2 \hat{J}(h_k)\|^2\right]}{2(L+\beta)} \\
 &\leq \left(L + \beta \right) \mathbb{E}[\|\Delta h_k\|^2] + \frac{\sigma_0}{\sqrt{N}} + \frac{\sigma_1^2}{2N(L+\beta)}
\end{align*}
Rearranging to isolate the expected squared step norm:
\begin{equation*}
 \left(L + \beta \right) \mathbb{E}[\|\Delta h_k\|^2] \geq \mathbb{E}[\|\nabla J(h_{k+1})\|] - \frac{\sigma_0}{\sqrt{N}} - \frac{\sigma_1^2}{2N(L+\beta)}
\end{equation*}
\begin{equation*}
 \mathbb{E}[\|h_{k+1}-h_k\|^2] \geq \frac{1}{L + \beta} \left( \mathbb{E}[\|\nabla J(h_{k+1})\|] - \frac{\sigma_0}{\sqrt{N}} - \frac{\sigma_1^2}{2N(L+\beta)} \right)
\end{equation*}

This completes the proof for the lower bound based on the gradient norm, using the corrected regularization term.

\subsection{Proof for convergence theorem}
\label{Convergence}
From Lemma \ref{upper bound}, the upper bound on the expected cubic step norm for $R \sim \text{Uniform}\{1, \ldots, M\}$ is:
\begin{equation*}
 \mathbb{E}\left[\|h_{R+1}-h_R\|^3\right] \leq \frac{36(J(h_1) - J^*)}{\beta M} + \frac{48\sqrt{3}}{\beta^{3/2}} \frac{\sigma_0^{3/2}}{N^{3/4}} + \frac{864}{\beta^3} \frac{\sigma_1^3}{N^{3/2}}.
\end{equation*}
As the number of iterations $M \to \infty$ and the batch size $N \to \infty$, the right-hand side approaches zero. Thus,
\begin{equation} \label{eq:cubic_norm_limit_R}
 \lim_{M, N \to \infty} \mathbb{E}\left[\|h_{R+1}-h_R\|^3\right] = 0.
\end{equation}
Using Lyapunov's inequality, $\mathbb{E}[\|h_{R+1}-h_R\|^2] \leq (\mathbb{E}[\|h_{R+1}-h_R\|^3])^{2/3}$. Taking the limit as $M, N \to \infty$ and using \eqref{eq:cubic_norm_limit_R}:
\begin{equation} \label{eq:squared_norm_limit_R}
 \lim_{M, N \to \infty} \mathbb{E}\left[\|h_{R+1}-h_R\|^2\right] = 0.
\end{equation}
From Lemma \ref{lower bound}, we rearrange the inequality which holds for any iteration $k$:
\begin{equation*}
 \mathbb{E}[\|\nabla J(h_{k+1})\|] \leq (L + \beta) \mathbb{E}[\|h_{k+1}-h_k\|^2] + \frac{\sigma_0}{\sqrt{N}} + \frac{\sigma_1^2}{2N(L+\beta)}.
\end{equation*}
Now, we take the expectation over the random index $R \sim \text{Uniform}\{1, \ldots, M\}$. Since $R$ selects one of the iterations $k \in \{1, \ldots, M\}$ uniformly, taking the expectation of the inequality with respect to $R$ effectively averages it:
\begin{equation*}
 \mathbb{E}_{R}\left[ \mathbb{E}[\|\nabla J(h_{R+1})\|] \right] \leq \mathbb{E}_{R}\left[ (L + \beta) \mathbb{E}[\|h_{R+1}-h_R\|^2] + \frac{\sigma_0}{\sqrt{N}} + \frac{\sigma_1^2}{2N(L+\beta)} \right].
\end{equation*}
Here, $\mathbb{E}_{R}$ denotes the expectation over the choice of $R$. Let $\mathbb{E}[\cdot]$ denote the total expectation (over the process history and $R$). The inequality becomes:
\begin{equation*}
 \mathbb{E}[\|\nabla J(h_{R+1})\|] \leq (L + \beta) \mathbb{E}[\|h_{R+1}-h_R\|^2] + \frac{\sigma_0}{\sqrt{N}} + \frac{\sigma_1^2}{2N(L+\beta)}.
\end{equation*}

Taking the limit as $M \to \infty$ and $N \to \infty$:
\begin{align*}
 \lim_{M, N \to \infty} \mathbb{E}[\|\nabla J(h_{R+1})\|] &\leq \lim_{M, N \to \infty} \left( (L + \beta) \mathbb{E}[\|h_{R+1}-h_R\|^2] + \frac{\sigma_0}{\sqrt{N}} + \frac{\sigma_1^2}{2N(L+\beta)} \right) \\
 &= (L + \beta) \lim_{M, N \to \infty} \mathbb{E}[\|h_{R+1}-h_R\|^2] + \lim_{N \to \infty} \frac{\sigma_0}{\sqrt{N}} + \lim_{N \to \infty} \frac{\sigma_1^2}{2N(L+\beta)} \\
 &= (L + \beta) \times 0 + 0 + 0 \quad \text{(Using \eqref{eq:squared_norm_limit_R})} \\
 &= 0.
\end{align*}
Since $\mathbb{E}[\|\nabla J(h_{R+1})\|] \ge 0$, we conclude that:
\begin{equation*}
 \lim_{M, N \to \infty} \mathbb{E}[\|\nabla J(h_{R+1})\|] = 0.
\end{equation*}
As $\mathbb{E}[\|\nabla J(h_R)\|]$ differs from $\mathbb{E}[\|\nabla J(h_{R+1})\|]$ by terms that vanish as $M \to \infty$ (typically $\frac{1}{M}(\mathbb{E}[\|\nabla J(h_{M+1})\|] - \mathbb{E}[\|\nabla J(h_1)\|])$), we can equivalently state:
\begin{equation*}
 \lim_{M, N \to \infty} \mathbb{E}[\|\nabla J(h_{R})\|] = 0.
\end{equation*}
This proves that the expected gradient norm at a randomly chosen iteration converges to zero.

\section{Proof for quadratic convergence}
\label{convergence_rate_proof}
Let the error at iteration $k$ be $e_k = h_k - h^*$. The update gives $h_{k+1} - h^* = h_k - h^* + \Delta h_k$, so $e_{k+1} = e_k + \Delta h_k$.
The update step satisfies the optimal condition \ref{Optimality}, which is given by:
 \begin{equation}\label{eq:deterministic_update_proof_nonum} \nonumber
  \nabla J(h_k) + \nabla^2 J(h_k) \circ \Delta h_k + \frac{\beta}{2} \|\Delta h_k\| \Delta h_k = 0.
 \end{equation}
Rearranging yields:
\begin{equation}\nonumber
 (\nabla^2 J(h_k) + \frac{\beta}{2} \|\Delta h_k\| \mathcal{I}) \circ \Delta h_k = - \nabla J(h_k)
\end{equation}
where $\mathcal{I}$ is the identity operator.

We expand $\nabla J(h_k)$ around $h^*$ using Taylor's theorem (similar to the derivation in Appendix \ref{Taylor upper bound}):
$$ \nabla J(h_k) = \nabla J(h^*) + \nabla^2 J(h^*) \circ (h_k - h^*) + \mathcal{R}_1(h_k, h^*) $$
where $\nabla J(h^*) = 0$ and the remainder term satisfies $\|\mathcal{R}_1(h_k, h^*)\| \leq C_1 \|h_k - h^*\|^2 = C_1 \|e_k\|^2$ for some constant $C_1$ when $h_k$ is near $h^*$.
Thus,
\begin{equation}\nonumber
 \nabla J(h_k) = \nabla^2 J(h^*) \circ e_k + \mathcal{R}_1(h_k, h^*)
\end{equation}
Substitute this into the rearranged update equation:
\begin{equation}\nonumber
 (\nabla^2 J(h_k) + \frac{\beta}{2} \|\Delta h_k\| \mathcal{I}) \circ \Delta h_k = - \nabla^2 J(h^*) \circ e_k - \mathcal{R}_1(h_k, h^*)
\end{equation}
Let $H_k = \nabla^2 J(h_k)$ and $H^* = \nabla^2 J(h^*)$. Let $A_k = H_k + \frac{\beta}{2} \|\Delta h_k\| \mathcal{I}$. The equation is $A_k \circ \Delta h_k = - H^* \circ e_k - \mathcal{R}_1(h_k, h^*)$.
Assuming the norm of the inverse operator $\|A_k^{-1}\|_{op}$ will be bounded by some constant $B$, and we assume that the update step is sufficiently small that $\|\Delta h_k\| \le K \|e_k\|$ for some $K > 0$.

Now, substitute $\Delta h_k = e_{k+1} - e_k$ into the equation $A_k \circ \Delta h_k = - H^* \circ e_k - \mathcal{R}_1(h_k, h^*)$:
\begin{align*}
 A_k \circ (e_{k+1} - e_k) &= - H^* \circ e_k - \mathcal{R}_1(h_k, h^*) \\
 A_k \circ e_{k+1} &= A_k \circ e_k - H^* \circ e_k - \mathcal{R}_1(h_k, h^*) \\
 A_k \circ e_{k+1} &= (H_k + \frac{\beta}{2} \|\Delta h_k\| \mathcal{I} - H^*) \circ e_k - \mathcal{R}_1(h_k, h^*) \\
 A_k \circ e_{k+1} &= (H_k - H^*) \circ e_k + \frac{\beta}{2} \|\Delta h_k\| e_k - \mathcal{R}_1(h_k, h^*)
\end{align*}
Applying the inverse $A_k^{-1}$:
\begin{equation}\nonumber
 e_{k+1} = A_k^{-1} \circ \left[ (H_k - H^*) \circ e_k + \frac{\beta}{2} \|\Delta h_k\| e_k - \mathcal{R}_1(h_k, h^*) \right]
\end{equation}
Taking norms and using the triangle inequality:
\begin{align*}
 \|e_{k+1}\| &\leq \|A_k^{-1}\|_{op} \left\| (H_k - H^*) \circ e_k + \frac{\beta}{2} \|\Delta h_k\| e_k - \mathcal{R}_1(h_k, h^*) \right\| \\
 &\leq \|A_k^{-1}\|_{op} \left( \|H_k - H^*\|_{op} \|e_k\| + \frac{\beta}{2} \|\Delta h_k\| \|e_k\| + \|\mathcal{R}_1(h_k, h^*)\| \right)
\end{align*}
Substitute the bounds: $\|A_k^{-1}\|_{op} \le B$, $\|H_k - H^*\|_{op} \le L \|e_k\|$, $\|\Delta h_k\| \le K \|e_k\|$, and $\|\mathcal{R}_1(h_k, h^*)\| \le C_1 \|e_k\|^2$.
\begin{align*}
 \|e_{k+1}\| &\leq B \left( (L \|e_k\|) \|e_k\| + \frac{\beta}{2} (K \|e_k\|) \|e_k\| + C_1 \|e_k\|^2 \right) \\
 &\leq B \left( L \|e_k\|^2 + \frac{\beta K}{2} \|e_k\|^2 + C_1 \|e_k\|^2 \right) \\
 &\leq B \left( L + \frac{\beta K}{2} + C_1 \right) \|e_k\|^2
\end{align*}
Setting $C_q = B (L + \frac{\beta K}{2} + C_1)$, which is a positive constant independent of $k$, we have shown that
\begin{equation}\nonumber
 \|h_{k+1} - h^*\| \leq C_q \|h_k - h^*\|^2
\end{equation}
This demonstrates local quadratic convergence for the deterministic version of the algorithm, provided $h_k$ is sufficiently close to $h^*$.

\section{Asset allocation experiment}
\label{asset_allocation}

In our investment planning MDP, we formulate a state-action framework that models investment decisions under varying market conditions and resource constraints. This model captures the fundamental trade-offs between risk and return across different market states while accounting for resource dynamics.

\textbf{The state} $s \in \mathcal{S}$ is characterized by a tuple $(r, m)$ where:
\begin{itemize}
    \item $r \in \{0,1,\ldots,R_{max}-1\}$ represents the discrete resource level, with $R_{max}$ being the maximum possible resource level
    \item $m \in \{0,1,2\}$ corresponds to market conditions (recession, stability, and prosperity, respectively)
\end{itemize}

The cardinality of the state space is $|\mathcal{S}| = R_{max} \times 3$.

\textbf{The action space} $\mathcal{A}$ comprises three distinct investment strategies:
\begin{itemize}
    \item $a=0$: Conservative investment (low risk/low return)
    \item $a=1$: Balanced investment (moderate risk/moderate return)
    \item $a=2$: Aggressive investment (high risk/high return)
\end{itemize}

\textbf{The state transition function} $P(s_{t+1} \mid s_{t}, a_{t})$ models the stochastic evolution of both resource levels and market conditions:

\begin{enumerate}
    \item Resource Dynamics: The probability of resource level transitions depends on the chosen action:
    \begin{itemize}
        \item Conservative strategy: $P(r_{t+1}|r_t,a_t=0) = [0.1, 0.8, 0.1, 0.0, 0.0]$ for $\Delta r \in \{-1,0,+1,+2,+3\}$
        \item Balanced strategy: $P(r_{t+1}|r_t,a_t=1) = [0.2, 0.2, 0.4, 0.2, 0.0]$ for $\Delta r \in \{-1,0,+1,+2,+3\}$
        \item Aggressive strategy: $P(r_{t+1}|r_t,a_t=2) = [0.4, 0.1, 0.1, 0.2, 0.2]$ for $\Delta r \in \{-1,0,+1,+2,+3\}$
    \end{itemize}

    \item Market Dynamics: Market state transitions follow a Markov chain with the following probabilities:
    \begin{itemize}
        \item Recession: $P(m_{t+1}|m_t=0) = [0.6, 0.3, 0.1]$ for $m_{t+1} \in \{0,1,2\}$
        \item Stability: $P(m_{t+1}|m_t=1) = [0.3, 0.4, 0.3]$ for $m_{t+1} \in \{0,1,2\}$
        \item Prosperity: $P(m_{t+1}|m_t=2) = [0.1, 0.3, 0.6]$ for $m_{t+1} \in \{0,1,2\}$
    \end{itemize}
\end{enumerate}

The joint transition probability is computed as:
\begin{align}
P((r_{t+1},m_{t+1})|(r_t,m_t),a_t) = P(r_{t+1}|r_t,a_t) \cdot P(m_{t+1}|m_t) \nonumber
\end{align}

\textbf{The reward function} $r(s_t, a_t)$ captures the expected immediate return for taking action $a_t$ in state $s_t=(r_t,m_t)$:

\begin{align}
r((r_t,m_t),a_t) = B(m_t,a_t) \cdot \frac{r_t+1}{R_{max}} \nonumber
\end{align}

where $B(m_t,a_t)$ is the base reward that depends on the market state and chosen action:

\begin{itemize}
    \item Conservative strategy $(a_t=0)$:
    \begin{itemize}
        \item Base reward $= 1.0$ for all market states, except
        \item Base reward $= 0.5$ in prosperity $(m_t=2)$ to represent opportunity cost
    \end{itemize}

    \item Balanced strategy $(a_t=1)$:
    \begin{itemize}
        \item Base reward $= 0.5$ in recession $(m_t=0)$
        \item Base reward $= 2.0$ in stability $(m_t=1)$
        \item Base reward $= 1.5$ in prosperity $(m_t=2)$
    \end{itemize}

    \item Aggressive strategy $(a_t=2)$:
    \begin{itemize}
        \item Base reward $= -1.0$ in recession $(m_t=0)$
        \item Base reward $= 1.0$ in stability $(m_t=1)$
        \item Base reward $= 3.0$ in prosperity $(m_t=2)$
    \end{itemize}
\end{itemize}

The resource scaling factor $\frac{r_t+1}{R_{max}}$ ensures that higher resource levels amplify rewards.

\textbf{The initial state distribution} $\rho(s_0)$ is typically set to start with a medium resource level and a randomly selected market state:

\begin{align}
\rho((r_0=\lfloor R_{max}/2 \rfloor, m_0=m)) = \frac{1}{3} \text{ for } m \in \{0,1,2\} \nonumber
\end{align}
In our experiment, $R_{max}$ is set as $5$, balancing sufficient environmental complexity with simplicity for visualization and optimal policy calculation.

\section{Optimization details}
\label{optimization_detail}

To solve the optimization problem formulated in Equation \eqref{optimization_final}, we implemented a conjugate gradient optimization framework based on the Newton-CG method. This approach combines the second-order convergence properties of Newton's method with the computational efficiency of the conjugate gradient algorithm, making it particularly suitable for our problem where the dimensionality of the Hessian matrix $H$ scales with the volume of trajectory data $N \times T$.

Specifically, we address the following optimization problem:
\begin{equation}
\nonumber
\boldsymbol{\bar{\alpha}}^{*} =\underset{\boldsymbol{\bar{\alpha}} \in \mathbb{R}^{NT}}{\operatorname{argmin}}\left\{\left\langle v, \boldsymbol{\bar{\alpha}}\right\rangle  +\frac{1}{2}\left\langle H \boldsymbol{\bar{\alpha}}, \boldsymbol{\bar{\alpha}}\right\rangle+\frac{\beta}{6}\left\|\boldsymbol{\bar{\alpha}}\right\|_2^{3}\right\}
\end{equation}

This optimization problem incorporates a linear term $\langle v, \boldsymbol{\bar{\alpha}}\rangle$, a quadratic term $\frac{1}{2}\langle H \boldsymbol{\bar{\alpha}}, \boldsymbol{\bar{\alpha}}\rangle$, and a cubic regularization term $\frac{\beta}{6}\|\boldsymbol{\bar{\alpha}}\|_2^{3}$. The objective function and its gradient are computed as:
\begin{align}
\nonumber
f(\boldsymbol{\bar{\alpha}}) &= \langle v, \boldsymbol{\bar{\alpha}}\rangle + \frac{1}{2}\langle H \boldsymbol{\bar{\alpha}}, \boldsymbol{\bar{\alpha}}\rangle + \frac{\beta}{6}\|\boldsymbol{\bar{\alpha}}\|_2^{3} \\
\nonumber
\nabla f(\boldsymbol{\bar{\alpha}}) &= v + H\boldsymbol{\bar{\alpha}} + \frac{\beta}{2}\|\boldsymbol{\bar{\alpha}}\|\boldsymbol{\bar{\alpha}}
\end{align}

In each Newton iteration, we determine the search direction by solving the linear system $(H + \frac{\beta}{2}\|\boldsymbol{\bar{\alpha}}\|\mathcal{I})\Delta\boldsymbol{\bar{\alpha}} = -\nabla f(\boldsymbol{\bar{\alpha}})$. The conjugate gradient method is employed to efficiently solve this linear system, avoiding the high computational cost of directly computing $(H + \frac{\beta}{2}\|\boldsymbol{\bar{\alpha}}\|\mathcal{I})^{-1}$. This method constructs a set of conjugate directions $\{p_i\}$ and progressively approximates the optimal solution through orthogonal projections. The algorithm proceeds as follows:

\begin{enumerate}
    \item Initialize residual $r_0 = -\nabla f(\boldsymbol{\bar{\alpha}}_0) = -(v + H\boldsymbol{\bar{\alpha}}_0 + \frac{\beta}{2}\|\boldsymbol{\bar{\alpha}}_0\|\boldsymbol{\bar{\alpha}}_0)$ and initial search direction $p_0 = r_0$
    \item For each iteration $k$:
    \begin{itemize}
        \item Compute optimal step size $\alpha_k = \frac{r_k^T r_k}{p_k^T (H + \frac{\beta}{2}\|\boldsymbol{\bar{\alpha}}_k\|\mathcal{I}) p_k}$
        \item Update solution $\Delta\boldsymbol{\bar{\alpha}}_{k+1} = \Delta\boldsymbol{\bar{\alpha}}_k + \alpha_k p_k$
        \item Update residual $r_{k+1} = r_k - \alpha_k (H + \frac{\beta}{2}\|\boldsymbol{\bar{\alpha}}_k\|\mathcal{I}) p_k$
        \item Calculate conjugate direction update coefficient $\beta_k = \frac{r_{k+1}^T r_{k+1}}{r_k^T r_k}$
        \item Update search direction $p_{k+1} = r_{k+1} + \beta_k p_k$
    \end{itemize}
\end{enumerate}

In our implementation, we utilized the \texttt{minimize} function from the SciPy optimization library, configured with the 'Newton-CG' method. To balance optimization accuracy and computational efficiency, we set the convergence tolerance to $10^{-3}$ and the maximum number of iterations to 500.

All experiments are conducted on an Intel(R) Xeon(R) Gold 5218 CPU operating at 2.30 GHz.

\section{Broader Impacts}
Our Policy Newton in RKHS algorithm offers several positive societal impacts, including improved computational efficiency in reinforcement learning training, which could reduce energy consumption and accelerate applications in resource allocation, healthcare planning, and financial systems. The enhanced sample efficiency could democratize access to sophisticated AI tools by requiring fewer computational resources.

\end{document}